\documentclass[10pt,twocolumn,letterpaper]{article}

\usepackage[pagenumbers]{cvpr} 
\usepackage{graphicx}
\usepackage{placeins} 
\usepackage{adjustbox}
\usepackage{float} 
\usepackage{subcaption}
\usepackage{caption}
\usepackage[table]{xcolor}
\newcommand{\rocketicon}{%
  \raisebox{-0.25em}{\includegraphics[height=2em]{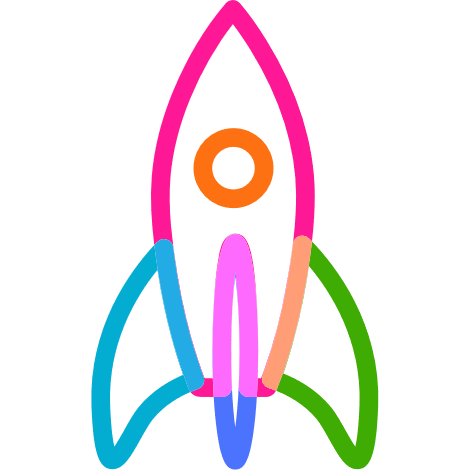}}%
}

\definecolor{cvprblue}{rgb}{0.21,0.49,0.74}
\usepackage[pagebackref,breaklinks,colorlinks,allcolors=cvprblue]{hyperref}

\def\paperID{*****} 
\def\confName{CVPR}
\def\confYear{2026}

\title{
\texorpdfstring{\rocketicon\ }{}No Cache Left Idle: Accelerating diffusion model via\\ Extreme-slimming Caching
}

\author{
{Tingyan Wen}$^{1\ast}$,
{Haoyu Li}$^{1\ast}$,
{Yihuang Chen}$^{2\dagger}$,
{Xing Zhou}$^{2}$,
{Lifei Zhu}$^{2}$,
{XueQian Wang}$^{1\dagger}$\\
$^1$Tsinghua University
$^2$Central Media Technology Institute, Huawei\\
{\tt\small \url{https://thu-accdiff.github.io/xslim-page/}}  
}

\begin{document}
\maketitle

\begin{abstract}
Diffusion models achieve remarkable generative quality, but computational overhead scales with step count, model depth, and sequence length. Feature caching is effective since adjacent timesteps yield highly similar features. However, an inherent trade-off remains: aggressive timestep reuse offers large speedups but can easily cross the critical line, hurting fidelity, while block- or token-level reuse is safer but yields limited computational savings. We present $\textbf{X-Slim}$ (e$\textbf{X}$treme-$\textbf{Slim}$ming Caching), a training-free, cache-based accelerator that, to our knowledge, is the first unified framework to exploit cacheable redundancy across timesteps, structure (blocks), and space (tokens). Rather than simply mixing levels, X-Slim adds a dual-threshold controller that turns caching into a $\textbf{push-then-polish}$ process. X-Slim first ``pushes'' reuse at the timestep level up to a early-warning line, then switches to lightweight block- and token-level refresh to ``polish'' the remaining redundancy---achieving no cache left idle. Once the critical line is crossed, full inference is required to reset accumulated error. At each level, context-aware indicators decide when and where to cache. Across diverse tasks, X-Slim advances the speed–quality frontier. On FLUX.1-dev and HunyuanVideo, it reduces latency by up to 4.97$\times$ and 3.52$\times$ with minimal perceptual loss. On DiT-XL/2, it reaches 3.13$\times$ acceleration with a lower FID of 2.42, outperforming prior state of the art.
\end{abstract}






\begingroup
\renewcommand\thefootnote{}%
\footnotetext{$^\ast$ Equal contribution. $^\dagger$ Corresponding author.}%
\addtocounter{footnote}{-1}%
\endgroup

\section{Introduction}
\label{sec:intro}

\begin{figure}[t]
  \centering
  \begin{adjustbox}{center,max width=\columnwidth}
    \includegraphics[width=\columnwidth]{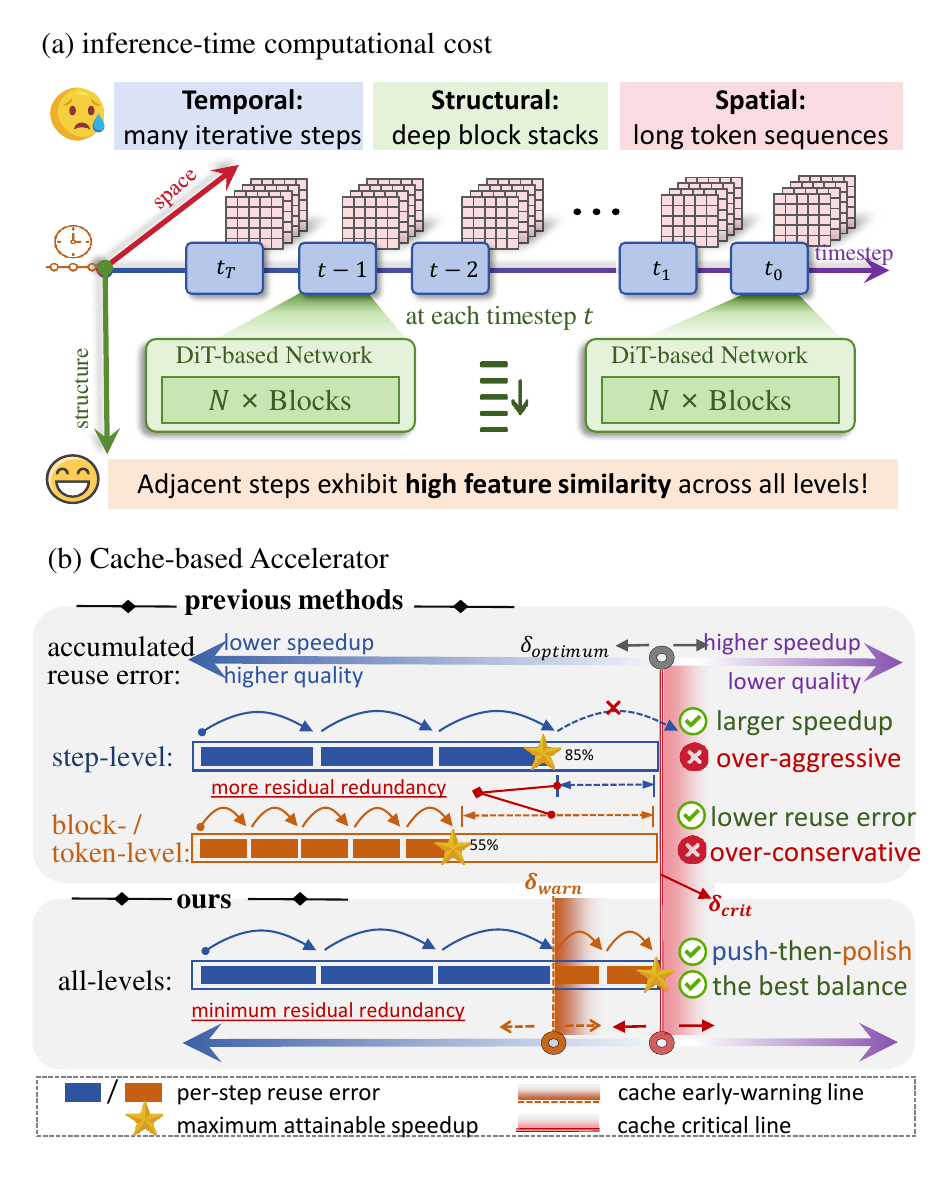} 
  \end{adjustbox}
  \vspace{-0.5cm}
  \caption{An overview of motivations.}
  \vspace{-0.6cm}
  \label{fig:motivation}
\end{figure}

Diffusion models\cite{dm-deep,dm-score,dm-ddpm} achieve remarkable generative quality across modalities\cite{sd,imagenvideo,diffwave,diffusion-lm}, but inference remains costly. As shown in Fig.~\ref{fig:motivation}(a), inference cost concentrates in three dimensions. Temporally, many iterative timesteps are required. Structurally, each step runs a deep stack of Transformer blocks\cite{dit} that increases compute with depth. Spatially, higher resolutions yield longer token sequences that raise attention and MLP cost.
Feature caching\cite{deepcache,fasterdiffusion,cacheme,teacache,taylorseers}, a training-free, plug-and-play acceleration method, has gained broad adoption for its effectiveness and simplicity. Features computed at one timestep are cached and reused at the next to avoid redundant computation. 
It motivates our first question—\emph{Is it possible to remove redundancy across temporal, structural, and spatial dimensions \textbf{via caching}?}

\maketitle
\begin{figure*}[!t]
  \centering
  \begin{adjustbox}{center,max width=\textwidth}
    \includegraphics[width=\textwidth]{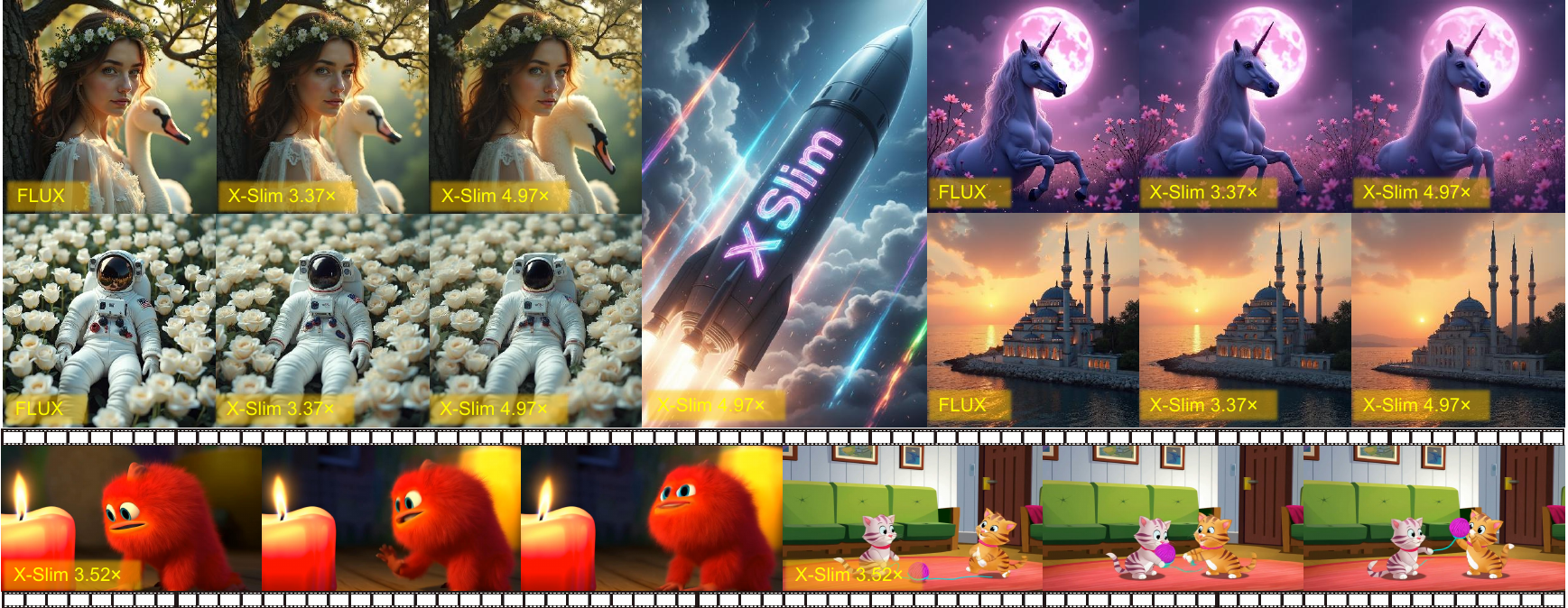}
  \end{adjustbox}
  \vspace{-0.5cm}
  \caption{Visualization of X-Slim on FLUX.1-dev (original, 3.37$\times$, 4.97$\times$) and on HunyuanVideo (3.52$\times$).}
  \vspace{-0.3cm}
  \label{fig:visualization}
  
\end{figure*}

Caching, however, comes in different forms. Most previous methods\cite{delta-dit,teacache} focus on a single level, which limits attainable speedups. In Fig.~\ref{fig:motivation}(b) we give a concrete example. Assume a maximum cache interval \(N=6\). The first step is computed in full, and features are reused for the next \(N-1\) steps.
Pure step-level\cite{teacache,taylorseers,easycache} reuse skips full forward passes and is fast. However, once adding one more cached step would push the cumulative reuse error beyond the threshold \(\delta_{\text{optimum}} \), the step-level schedule cannot extend further, leaving safe reuse opportunities unused. Pure block/token-level\cite{delta-dit,sortblock,dato,toca} updates recompute only parts of the network, so quality is stable but speed gains accumulate slowly. Under the same maximum error threshold, both leave residual redundancy and limit speedup.
This leads to our second key question: \emph{How can we \textbf{harness the benefits} of aggressive step-level and conservative block/token reuse to achieve maximal speed while preserving quality?}

To address these two questions, we propose \textbf{X-Slim} (e\textbf{X}treme-\textbf{Slim}ming Caching), a training-free, cache-based accelerator that exploits cacheable redundancy across steps, structure, and space.

{\bf Push-then-Polish Caching:}
Directly mixing multi-level reuse creates a large per-step search space \cite{dato,duca,astraea} with hand-crafted, constraint-heavy policies that are slow to tune and brittle. Instead, X\text{-}Slim introduces a dual-threshold controller with a clear push-then-polish caching paradigm. The controller sets an early-warning line \( \delta_{\text{warn}} \) that signals when aggressive step-level skipping leaves the safe region. At this point, a partial refresh suffices while a full inference is unnecessary. A critical line \( \delta_{\text{crit}} \) triggers a full inference step to reset accumulated error. X\text{-}Slim first \emph{pushes} at the timestep level to gain speed quickly, then \emph{polishes} with lightweight refresh. 
This simple paradigm is highly effective. It dramatically reduces the policy search space compared with direct multi-level mixing while avoiding complex manual tuning and ad hoc constraints.

\begin{figure}[t] 
  \centering
  \begin{subfigure}[b]{0.49\linewidth} 
      \centering
      \includegraphics[width=\linewidth]{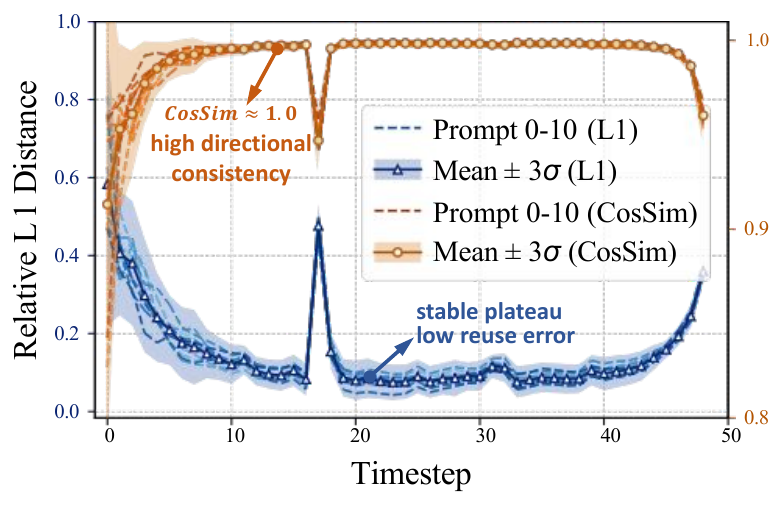} 
      \vspace{-0.6cm}
      \caption{step-level (random prompts)} 
      \label{fig:step_level}
  \end{subfigure}
  \hfill
  \begin{subfigure}[b]{0.49\linewidth} %
      \centering
      \includegraphics[width=\linewidth]{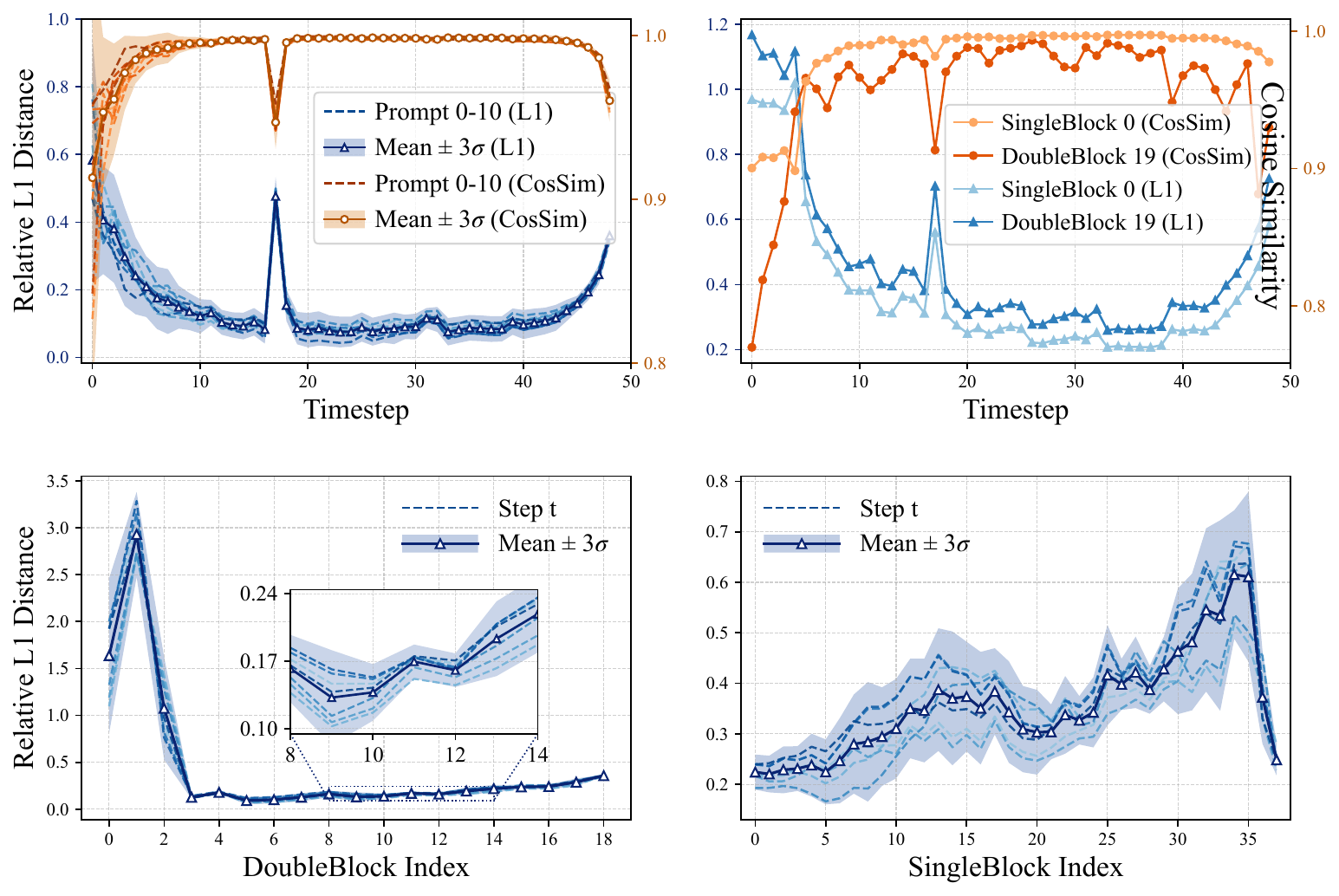}
      \vspace{-0.6cm}
      \caption{block-level (default prompt)}
      \label{fig:block_level}
  \end{subfigure}
  
  
  \begin{subfigure}[b]{0.49\linewidth}
      \centering
      \includegraphics[width=\linewidth]{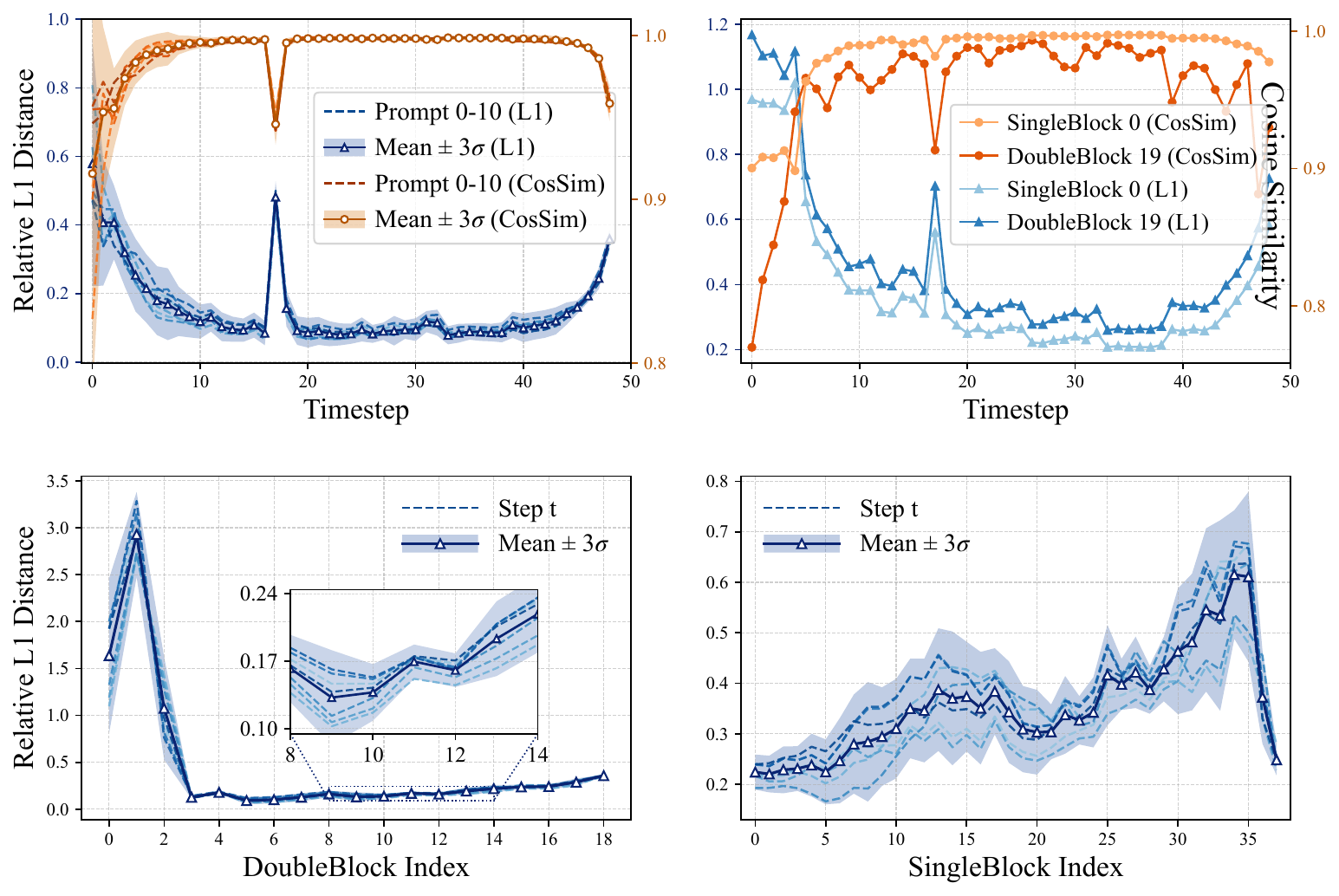}
      \vspace{-0.6cm}
      \caption{doubleblock effect per step}
      \label{fig:doubleblock}
  \end{subfigure}
  \hfill
  \begin{subfigure}[b]{0.49\linewidth}
      \centering
      \includegraphics[width=\linewidth]{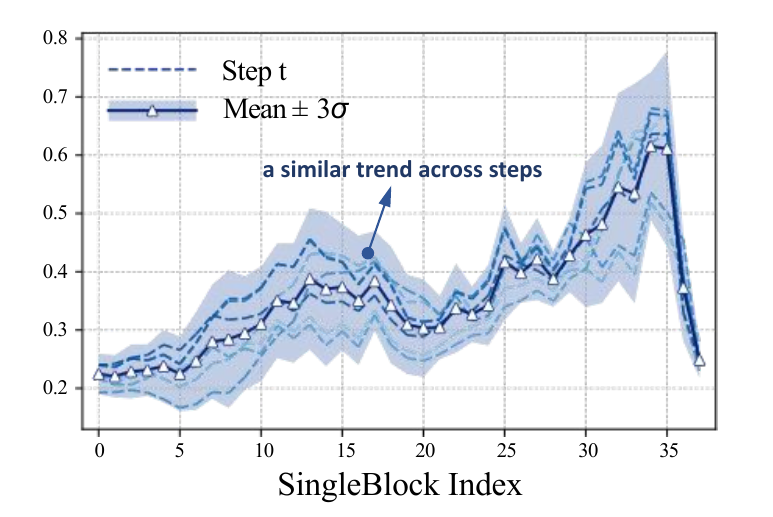}
      \vspace{-0.6cm}
      \caption{singleblock effect per step}
      \label{fig:singleblock}
  \end{subfigure}

  \caption{Analysis of feature reuse. We report two metrics, L1-relative distance for reuse-error magnitude and cosine similarity for directional consistency. In (a), ten random prompts are from DrawBench~\cite{drawbench}.}
  \vspace{-0.5cm}
  \label{fig:four-panel-analysis}
\end{figure}

{\bf Level-specific Strategy:} We observe that each level exhibits distinct reuse dynamics. At the step level, we visualize per-timestep relative-change curves on diverse prompt benchmarks \cite{drawbench,hps} and find that these curves are consistently U-shaped and weakly prompt-dependent (see Fig.~\ref{fig:step_level} and Supplementary Section~\ref{sec:diff-prompts}). 
Moreover, existing dynamic step schedulers \cite{teacache,easycache} converge to similar strategies but require heavy preprocessing.
Therefore, temporal caching effectiveness is not driven by prompts but is governed primarily by the model's intrinsic denoising dynamics.
To further support this claim, we apply caching in three temporal phases, early, middle, and late. We observe that early and late caching degrade structure and detail, while middle caching is most robust (see Supplementary Section~\ref{sec:position-sensitivity}).
For blocks, sensitivity varies with depth yet the depth-wise pattern is consistent across timesteps, see Figs.~\ref{fig:doubleblock} and \ref{fig:singleblock}. Tokens are strongly content dependent. Building on these observations, X-Slim adopts a hybrid, level-specific selection strategy that plays to each level's strengths. At the step level, we use a static schedule on a small batch, which is simple, efficient, model-friendly, and avoids complex preprocessing. At the block and token levels, we adopt a dynamic, context-aware policy. Each level is governed by its own selection function, providing precise per-level control with minimal overhead.


Our main contributions are summarized as follows:

\begin{itemize}
\item We propose \textbf{X-Slim}, the first unified framework to jointly exploit caching over timesteps, structure, and space, fully tapping cacheable computation for \emph{extreme slimming}.
\item We introduce a \emph{push-then-polish} caching paradigm with a \emph{level-specific strategy}. Through a dual-threshold controller and a hybrid scheme, X\text{-}Slim provides precise per-level control with minimal overhead.
\item Training-free and plug-and-play, X\text{-}Slim surpasses prior methods with up to \textbf{4.97$\times$} and \textbf{3.52$\times$} lower latency (text-to-image and text-to-video), and \textbf{3.13$\times$} on DiT-XL/2 with a \textbf{2.42} FID improvement.

\end{itemize}

\section{Related Work}
\label{sec:related work}
\subsection{Efficient Diffusion Models}

Diffusion models produce high-fidelity images\cite{sdxl,sd3} and videos\cite{imagenvideo,videofusion}, but inference remains costly. Existing accelerations follow two complementary tracks. Step reduction replaces long trajectories with shorter ones using improved samplers (e.g., DDIM\cite{ddim}, DPM-Solver\cite{dpmsolver}) or distillation\cite{guideddistillation,dmd} that compresses multi-step teachers into few- or single-step students. Per-step cost reduction lightens the network with pruning\cite{structural_pruning_diffusion, bksdm}, quantization\cite{quantization1,quantization2,quantization3}, and token-level shrinking\cite{tofu} to cut FLOPs within each step. Although effective, they require extra training/fine-tuning or architectural changes to maintain fidelity, reducing portability across backbones and tasks.


\subsection{Diffusion Model Caching}

Caching has recently become a promising accelerator by reusing intermediate features across timesteps. Early U-Net–based\cite{unet} methods such as DeepCache\cite{deepcache} and Faster Diffusion\cite{fasterdiffusion} pioneered feature reuse across adjacent steps. With the emergence of Diffusion Transformers (DiTs)\cite{dit},  FORA\cite{fora} and $\Delta$-DiT\cite{delta-dit} extend caching to transformer backbones.  At the step level, TeaCache\cite{teacache} triggers reuse decisions based on timestep-embedding differences. Predictive approaches like TaylorSeer\cite{taylorseers} and AB-Cache\cite{abcache} replace direct reuse with feature forecasting to stabilize long-interval caching. Methods such as $\Delta$-DiT\cite{delta-dit}, SortBlock\cite{sortblock}, and BACache\cite{bacache} analyze and exploit block-level redundancy and scheduling, whereas TokenCache\cite{tokencache}, ToCa\cite{toca}, and DaTo\cite{dato} operate at the token level. However, single-axis methods achieve only \textbf{local} optima in diffusion acceleration.

\section{Method}
\label{sec:method}

\maketitle

\begin{figure*}[!t]
  \centering
  \begin{adjustbox}{center,max width=\textwidth}
    \includegraphics[width=1\textwidth]{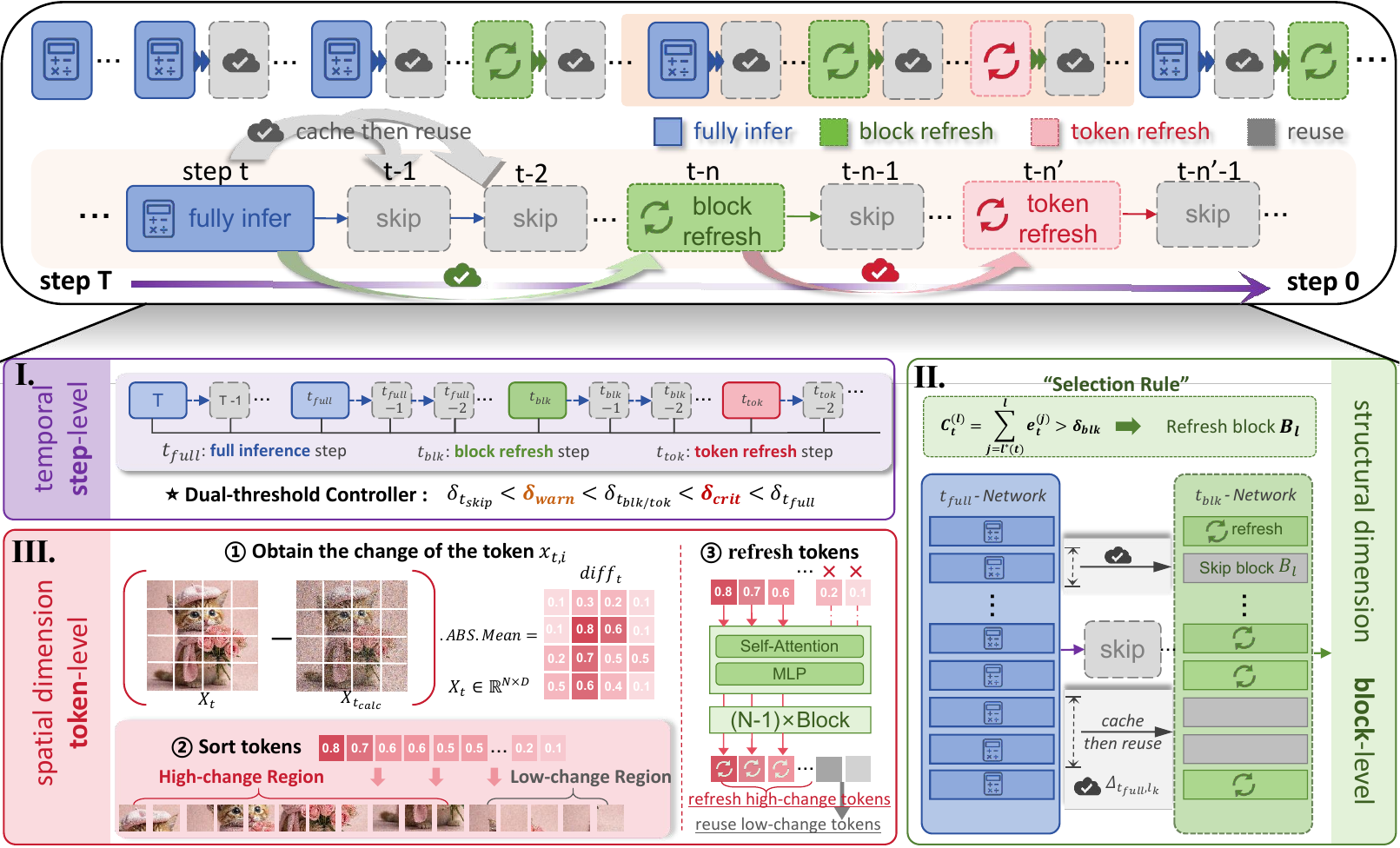}
  \end{adjustbox}
  \vspace{-0.5cm}
  \caption{Overview of X\text{-}Slim. \textbf{Timeline (top)} shows our push-then-polish process across timesteps. Reverse diffusion interleaves skip, full inference, block/token refresh steps. In skip or refresh steps, resued values come from the most recent computed step. \textbf{Level\mbox{-}specific strategy:} \textbf{I. Temporal level.} A dual\mbox{-}threshold controller forms a coarse\mbox{-}to\mbox{-}fine schedule. \textbf{II. Structural level.} Blocks receive an effect score from their input\mbox{-}output change. The block is refreshed when the accumulated effect score exceeds \( \delta_{\text{blk}} \), otherwise it reuses its cache. 
  \textbf{III. Spatial level.} Rank tokens by per-token change. Refresh high-change tokens and reuse low-change tokens from cache. }

  \vspace{-0.3cm}
  \label{fig:overview}
\end{figure*}


\subsection{Preliminary}

{\bf Denoising Diffusion Models.} Diffusion models consist of a forward noising process and a reverse denoising process. The forward process adds Gaussian noise to \(x_0\sim p(x_0)\) over \(T\) steps,
\[
q(x_t\mid x_{t-1})=\mathcal{N}\!\big(x_t;\,\sqrt{1-\beta_t}\,x_{t-1},\,\beta_t I\big),
\]
where \(\beta_t\in(0,1)\) is the noise schedule. After many steps \(x_T\approx\mathcal{N}(0,I)\). The reverse process predicts \(x_{t-1}\) from \(x_t\),
\[
p_\theta(x_{t-1}\mid x_t)=\mathcal{N}\!\big(x_{t-1};\,\mu_\theta(x_t,t),\,\Sigma_\theta(x_t,t)\big),
\]
with \(\mu_\theta\) and \(\Sigma_\theta\) parameterized by a neural network.

\paragraph{Transformer-based backbone.}
A DiT denoiser is a stack of \(L\) Transformer blocks,
\(\mathrm{Net}=B_{1}\circ B_{2}\circ \cdots \circ B_{L}\),
where each block \(B_l\) contains multi-head self-attention, an MLP, and conditioning via AdaLN.
At each timestep the input is a token sequence
\(\mathcal{X}=\{x_1,x_2,\ldots,x_{N}\}\in\mathbb{R}^{N\times D}\),
where \(N\) is the sequence length and \(D\) is the hidden dimension.
Overall compute and latency scale with timesteps \(T\), blocks \(L\), and tokens \(N\).
Increasing any of these factors raises computational complexity and inference latency.

\subsection{Push-then-Polish Caching}
Rather than a direct mixture, X-Slim introduces a push-then-polish caching paradigm, exploiting cachable redundancy across temporal, structural, and spatial dimensions. 

\paragraph{Cache-based Metric.}
Adjacent timesteps are highly similar during denoising, enabling reuse of previously computed features to reduce computation.
Let \(I_t\) and \(O_t\) denote the input and output at step \(t\) (at any level: step, block, or token), and define the change \(\Delta_t = O_t - I_t\).
With caching, we approximate the current output by reusing the last change: \(O_t \approx I_t + \Delta_{t-1}\).
To monitor drift, we accumulate a reuse error from the most recent computed step \(t_{\text{calc}}\), which may be a full inference or a partial refresh, to the current step \(t\):
\[
\mathrm{err}(t) \;=\; \sum_{k=t_{\text{calc}}+1}^{t}
\frac{\bigl\| \Delta_k - \Delta_{k-1} \bigr\|_{1}}{\bigl\| \Delta_{k-1} \bigr\|_{1}} \, .
\]

\paragraph{Dual-threshold Controller.}
Based on the cumulative reuse error, X-Slim determines how to process each timestep using two thresholds, \(\delta_{\text{warn}}\) and \(\delta_{\text{crit}}\). These thresholds divide timesteps into three modes: \textbf{skip step}, \textbf{full inference step}, and \textbf{refresh step}, forming a push-then-polish adaptive caching paradigm.

\textbf{Skip step.}
When the reuse error is less than \(\delta_{\text{warn}}\), the model is in a safe and stable region where the change between timesteps is minor. The current step directly reuses the previous result without any computation, and the output is updated as \(O_t \simeq I_t + \Delta_{t-1}\). This coarse, step-level skipping provides the highest acceleration without any extra cost.

\textbf{Full inference step.}
When the reuse error exceeds \(\delta_{\text{crit}}\), full computation is required to restore accuracy. The model performs a complete forward pass to obtain \(O_t\), refreshes all cached features, and resets the accumulated error. 

\textbf{Refresh step.}
When the reuse error lies between the two thresholds, i.e., \(\delta_{\text{warn}} < \text{reuse error} < \delta_{\text{crit}}\), the model enters the refresh step. At this stage, some error has accumulated and needs correction. A full recomputation is wasteful since cacheable redundancy may still remain. To address this, X-Slim performs a partial refresh that only recomputes the necessary part and keep those stable fine-grained layers/tokens as reused, exploiting still-cachable redundancy.
The operation is formulated as
\[
O_t = \text{Compute}(S_t; I_t) \oplus \text{Reuse}(\overline{S_{t-1}}),
\]
where \(S_t\) denotes the selected subset of layers or tokens, and \(\oplus\) represents the combination of recomputed and reused parts. 
This corrects accumulated errors with minimal additional cost, achieving fine-grained control between skipping and full inference.

In this design, \(\delta_{\text{warn}}\) acts as a cache early-warning line, indicating when to leave the safe skip region and introduce partial computation to correct potential reuse errors, while \(\delta_{\text{crit}}\) defines the cache critical line that triggers full inference to restore accuracy. If  \(\delta_{\text{warn}} = \delta_{\text{crit}}\), the strategy reduces to an aggressive step-level cache. If \(\delta_{\text{warn}} = 0\), it becomes a conservative fine-grained cache. By setting these thresholds, X-Slim helps avoid leaving cacheable redundancy idle while preserving quality.

\subsection{Level-specific Strategy}
Different levels exhibit distinct reuse dynamics. Adjacent steps follow a U-shaped pattern and are weakly prompt dependent. Block sensitivity varies with depth yet follows a consistent depth-wise pattern across timesteps. Tokens are largely content dependent. Building on these properties, X-Slim adopts a hybrid level-specific strategy that plays to each level's strengths.

\paragraph{Temporal level.}
We calibrate a static step-level schedule on a small calibration set, guided by the model's inherent dynamics.
At runtime, X\text{-}Slim follows a coarse-to-fine schedule: skip, refresh and full inference. During refresh, block- and token-level updates are interleaved to correct errors efficiently, after which the controller re-evaluates the error and resumes the schedule. 

\paragraph{Structural level.}
We define a block's effect as its input–output relative change
\[
e_t^{(l)} \;=\; \frac{\bigl\|O_t^{(l)} - I_t^{(l)}\bigr\|_{1}}{\bigl\|I_t^{(l)}\bigr\|_{1}} .
\]
Larger \(e_t^{(l)}\) indicates higher influence and greater risk under caching.
Empirically, sensitivity varies across depth, yet the depth-wise pattern is consistent between adjacent timesteps.
Leveraging this property, we dynamically filter redundant blocks at full reference steps by accumulating effects over the current skipped run and refreshing only when the accumulated importance exceeds a threshold:
\[
\text{refresh block } l \;\iff\;
C_t^{(l)} \;:=\; \sum_{j=l^\star(t)}^{\,l} e_t^{(j)} \;>\; \delta_{\text{blk}},
\]
where \(l^\star(t)\) denotes the start index of the current skipped segment and is reset after each refresh.
If refreshed, we recompute block \(l\), reset the cumulative value, advance the run start to \(l^\star(t)\!\leftarrow l+1\), and continue. Otherwise, the block reuses its cached \(\Delta_{l}\).

\paragraph{Spatial level.}
X\text{-}Slim adopts a \emph{context-aware} policy that dynamically selects tokens in low/high-change regions for reuse or refresh.
Let \(X_t\in\mathbb{R}^{N\times D}\) denote the hidden-state input at timestep \(t\).
For token \(i\), define the absolute change with respect to the most recent computed step \(t_{\text{calc}}\) as
\[
\mathrm{diff}_i(t)=\frac{1}{D}\sum_{d=1}^{D}\bigl|x_{i,t,d}-x_{i,t_{\text{calc}},d}\bigr| .
\]
We sort tokens by \(\mathrm{diff}_i(t)\) and reuse the low-change tokens as residual redundancy, while recomputing the remaining tokens through the network.
This concentrates computation on high-change regions and safely skips low-change tokens.

\begin{figure}[!t]
  \centering

  \begin{subfigure}{\linewidth}
    \centering
    \includegraphics[width=\linewidth]{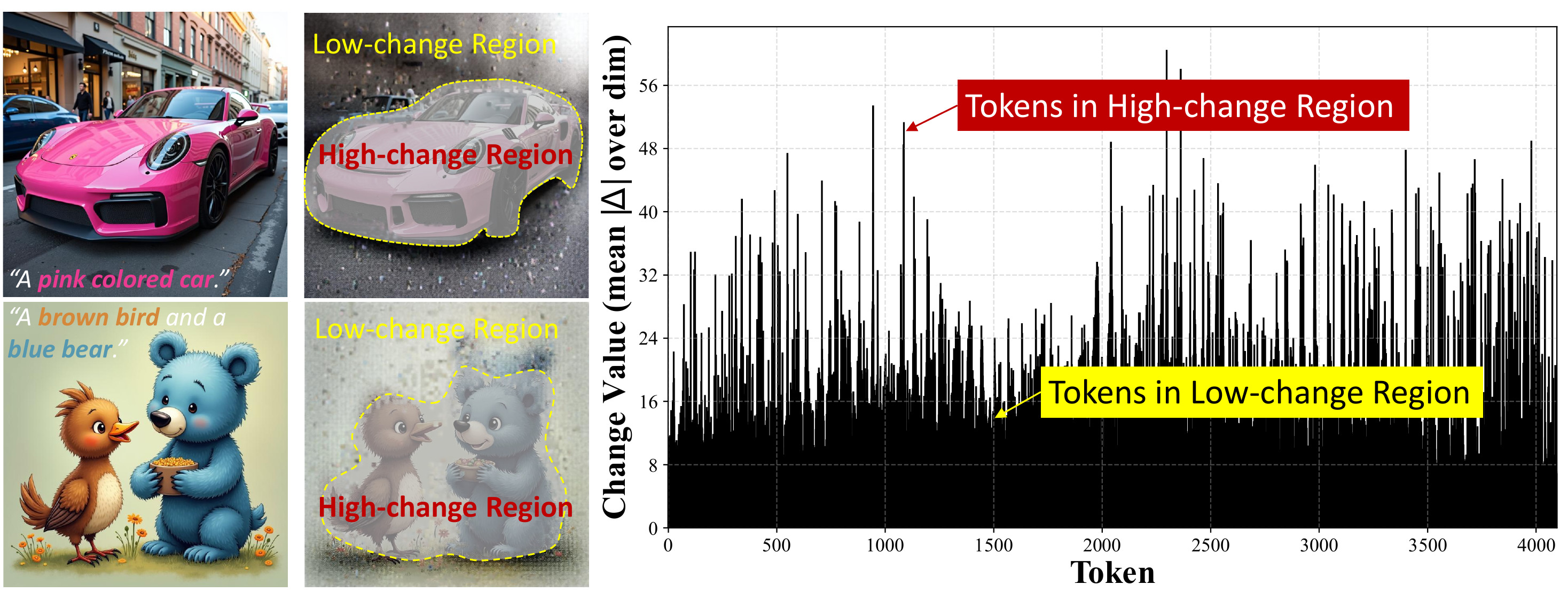}
    \caption{Token Selection: high- and low-change tokens.}
    \label{fig:tokenchoice}
  \end{subfigure}

  \vspace{1mm} 

  \begin{subfigure}{\linewidth}
    \centering
    \includegraphics[width=\linewidth]{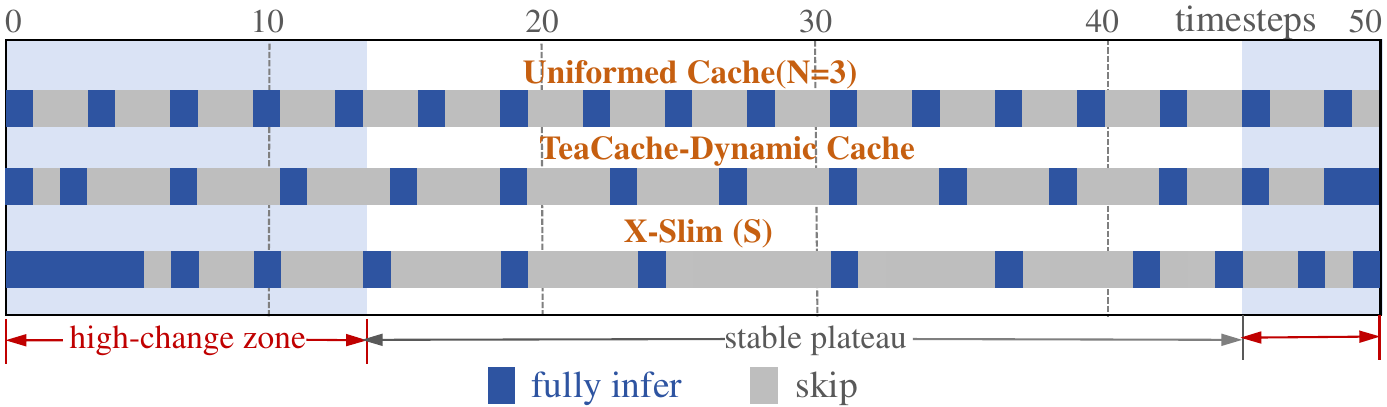}
    \caption{Comparison of step-level strategy across methods. } 
    \label{fig:step-strategy}
  \end{subfigure}
  \vspace{-0.5cm}
  \caption{Selection visualization on FLUX.1-dev.}
  \vspace{-0.3cm}
  \label{fig:selec-visual}
  \vspace{-2mm} 
\end{figure}

\paragraph{Selection Visualization.}
In Fig.~\ref{fig:tokenchoice}, we visualize high and low change tokens between adjacent steps. Updating only high change tokens concentrates compute on the subject such as “pink colored car,” while the background receives fewer updates. At the timestep level the model's intrinsic denoising dynamics dominate. Early caching harms global structure. Late caching blurs details and leaves residual noise. Mid phase caching is consistently robust. See Supplementary Section~\ref{sec:position-sensitivity} for more analysis. To substantiate this, we compare our step-only variant X\text{-}Slim(S) with step level baselines Uniform Cache and TeaCache in Fig.~\ref{fig:step-strategy}. X\text{-}Slim(S) follows the model's intrinsic denoising dynamics more closely and achieves the best performance (see Tab.~\ref{tab_flux}).


\section{Experiments}
\label{sec:experiment}

\begin{table*}[htbp]
\centering
\caption{\textbf{Quantitative comparison in text-to-image generation }on DrawBench\cite{drawbench} with FLUX.1-dev}
\vspace{-0.2cm} 
\label{tab_flux}
\setlength\tabcolsep{3pt}  
\renewcommand{\arraystretch}{0.9} 

\begin{tabular}{l|cc|ccccc} 
\toprule
\textbf{Method}
 & \multicolumn{2}{c|}{\textbf{Efficiency}} 
 & \multicolumn{5}{c}{\textbf{Visual Quality}} \\ 
\textbf{FLUX.1-dev}
 & \textbf{Latency(s)$\downarrow$} & \textbf{Speed$\uparrow$}
 & \textbf{ImageReward\cite{imagereward}$\uparrow$} & \textbf{CLIP\cite{clip}$\uparrow$} 
 & \textbf{PSNR\cite{iqa}$\uparrow$} & \textbf{SSIM\cite{iqa}$\uparrow$} & \textbf{LPIPS\cite{lpips}$\downarrow$} \\
\midrule
\textbf{Original: 50 steps}              & 27.32 & 1.00$\times$ & 0.9886 {\scriptsize \textcolor{gray}{(+0.0000)}} & 0.3159 & -- & -- & -- \\
$50\%$: 25 steps                        & 13.78 & 1.98$\times$ & 0.9640 {\scriptsize \textcolor{gray}{(-0.0246)}} & 0.3152 & 17.922 & 0.7399 & 0.2867 \\
\midrule
$\mathbf{30\%}$\textbf{: 15 steps}      &  8.40 & 3.25$\times$ & 0.9016 {\scriptsize \textcolor{gray}{(-0.0870)}}& 0.3153 & 15.494 & 0.6585 & 0.4035 \\
Uniform Cache ($\mathcal{N}=3$)         &  9.58 & 2.85$\times$ & 0.8959 {\scriptsize \textcolor{gray}{(-0.0927)}}& 0.3150 & 15.816 & 0.6726 & 0.3828 \\
$\Delta$-Dit ($\mathcal{N}=3$)          & 12.36 & 2.21$\times$ & 0.8987 {\scriptsize \textcolor{gray}{(-0.0899)}}& 0.3151 & 15.633 & 0.6649 & 0.3915 \\
ToCa ($\mathcal{N}=5$)                  &  8.73 & 3.13$\times$ & 0.8951 {\scriptsize \textcolor{gray}{(-0.0935)}}& 0.3105 & 17.288 & 0.6007 & 0.4217 \\
TaylorSeer ($\mathcal{N}=3$)            & 11.63 & 2.35$\times$ & 0.9814 {\scriptsize \textcolor{gray}{(-0.0072)}}& 0.3158 & 19.603 & 0.7820 & 0.2231 \\
TeaCache ($\delta=0.6$)                 &  8.42 & 3.24$\times$ & 0.9718 {\scriptsize \textcolor{gray}{(-0.0168)}}& 0.3145 & 16.938 & 0.7006 & 0.3433 \\
\rowcolor{gray!20}
\textbf{X-Slim(S)-slow }                 &  9.08 & 3.01$\times$ & \underline{0.9901} {\scriptsize \textcolor{blue}{(+0.0015)}} & \textbf{0.3164} & \textbf{21.717} & \textbf{0.8162} & \textbf{0.1905} \\
\rowcolor{gray!20}
\textbf{X-Slim(C2F)-slow}                & 8.11  & 3.37$\times$   & \textbf{0.9902} {\scriptsize \textcolor{blue}{(+0.0016)}}& \underline{0.3162} & \underline{21.300} & \underline{0.8097} & \underline{0.2105} \\
\midrule
$\mathbf{20\%}$\textbf{: 10 steps}      &  5.74 & 4.75$\times$ & 0.7878 {\scriptsize \textcolor{gray}{(-0.2008)}}& 0.3139 & 14.529 & 0.6153 & 0.4733 \\
Uniform Cache ($\mathcal{N}=5$)         &  5.80 & 4.71$\times$ & 0.7991 {\scriptsize \textcolor{gray}{(-0.1895)}}& 0.3142 & 14.530 & 0.6120 & 0.4755 \\
$\Delta$-Dit ($\mathcal{N}=5$)            &  7.51 & 3.64$\times$ & 0.8234 {\scriptsize \textcolor{gray}{(-0.1652)}}& 0.3019 & 14.103 & 0.6391 & 0.4588 \\
ToCa ($\mathcal{N}=8$)                  &  7.19 & 3.80$\times$ & 0.8608 {\scriptsize \textcolor{gray}{(-0.1278)}}& 0.3039 & 16.351 & 0.5243 & 0.5319 \\
TaylorSeer ($\mathcal{N}=6$)            &  7.59 & 3.60$\times$ & 0.9688 {\scriptsize \textcolor{gray}{(-0.0198)}}& \textbf{0.3182} & 15.780 & 0.6514 & 0.4014 \\
TeaCache ($\delta=1.0$)                 &  5.96 & 4.58$\times$ & 0.8619 {\scriptsize \textcolor{gray}{(-0.1267)}}& 0.3137 & 15.472 & 0.6437 & 0.4378 \\
\rowcolor{gray!20}
\textbf{X-Slim(S)-fast}                  &  5.83 & 4.69$\times$ & \textbf{0.9810} {\scriptsize \textcolor{blue}{(-0.0076)}}& \underline{0.3156} & \textbf{19.560} & \textbf{0.7520} & \textbf{0.2884} \\
\rowcolor{gray!20}
\textbf{X-Slim(C2F)-fast}                & 5.50  & 4.97$\times$ & \underline{0.9806} {\scriptsize \textcolor{blue}{(-0.0080)}}& 0.3153 & \underline{19.241} & \underline{0.7321} & \underline{0.2919} \\
\bottomrule
\end{tabular}
\footnotesize
\begin{itemize}
    \item Note that the acceleration ratio is calculated based on the actual latency measured on the same device, which is lower than theoretical FLOPs-based speedups but represents the real-world performance.
\end{itemize}
\vspace{-0.4cm}
\end{table*}

\subsection{Settings}
\paragraph{Base Models and Compared Methods.}s
We evaluate our method on three representative diffusion backbones and tasks: 
the text-to-image model FLUX.1-dev\cite{flux2024}, 
the text-to-video model HunyuanVideo\cite{hunyuanvideo}, 
and the class-conditional image generation model DiT-XL/2\cite{dit}. 
For comparison, we include recent cache-based accelerators covering step-, block-, and token-level designs: 
$\Delta$-DiT\cite{delta-dit}, ToCa\cite{toca}, TeaCache\cite{teacache}, TaylorSeer\cite{taylorseers}, EasyCache\cite{easycache}, 
as well as a fixed-interval caching schedule Uniform Cache and the full model without caching.

\paragraph{Configurations.}
We report two configurations of our method. \textbf{X\text{-}Slim(S)} represents our step-only variant and is compared with other step-level methods. \textbf{X\text{-}Slim (coarse to fine, C2F)} further strengthens the variant by adding fine grained reuse at the structural and spatial levels, removing remaining redundancy for additional acceleration with minimal quality loss.
Moreover, for each task, we provide two runtime settings, named \textbf{slow} and \textbf{fast}, to balance speed and quality.
Further details are provided in Supplementary Section~\ref{sec:detail-exp-settings}.
\begin{figure}[!t]
  \centering
  \includegraphics[width=\linewidth]{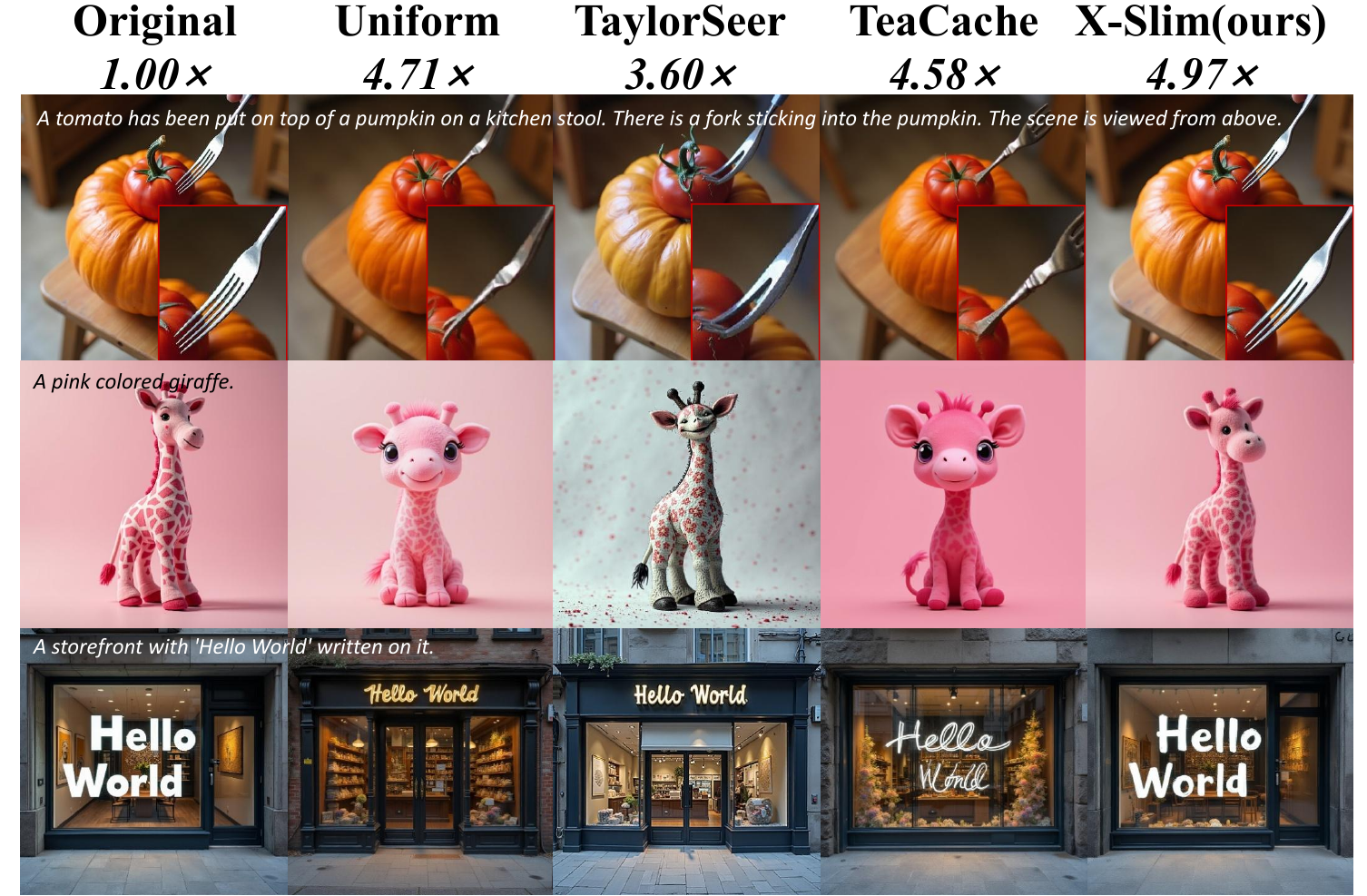}
  \caption{Qualitative comparison on FLUX.1-dev.}
  \vspace{-0.6cm}
  \label{fig:compar-flux}
\end{figure}
\subsection{Main Results.}
\paragraph{Quantitative Comparison.}

{\bf{Text-to-Image Generation.}} Tab.~\ref{tab_flux} summarizes the results on FLUX.1-dev for text-to-image task. 
In the slow setting, TeaCache and TaylorSeer deliver approximately 3$\times$ speedups with ImageReward\cite{imagereward} scores of 0.9718 and 0.9814, respectively. Our coarse-grained X-Slim(S) attains a comparable speedup with a higher score of 0.9901. With fine-grained block- and token-level caching, X-Slim(C2F) raises the speedup to 3.37$\times$ while maintaining nearly the same quality. When pushed to higher acceleration, baselines degrade sharply. By contrast, X-Slim(S) reaches 4.69$\times$ with only minor quality loss, and X-Slim(C2F) achieves 4.97$\times$ with essentially no additional degradation, establishing a new state of the art.



\begin{table}[htbp]
\centering
\caption{\textbf{Quantitative comparison in class-to-image generation }on ImageNet\cite{imagenet} with DiT-XL/2.}
\label{tab_dit}
\vspace{-0.2cm} 
\setlength{\tabcolsep}{1pt}
\small
\renewcommand{\arraystretch}{0.9} 

\resizebox{\columnwidth}{!}{
\begin{tabular}{l|cc|ccc}
    \toprule
    \textbf{Method}
    & \multicolumn{2}{c|}{\textbf{Efficiency}}
    & \multicolumn{3}{c}{\textbf{Visual Quality}} \\
    \textbf{DiT-XL/2}
    & \textbf{Latency (s)$\downarrow$} & \textbf{Speed$\uparrow$}
    & \textbf{FID\cite{fid}$\downarrow$}     & \textbf{sFID\cite{fid}$\downarrow$} & \textbf{IS\cite{is}$\uparrow$} \\
    \midrule
    \textbf{DDIM-50 steps†}          & 1.564 & 1.00$\times$ & 2.32 & 4.32 & 241.25 \\
    $\Delta$-DiT ($\mathcal{N}=2$)†  & 0.875 & 1.79$\times$ & 2.69 & 4.67 & 225.99 \\
    $\Delta$-DiT ($\mathcal{N}=3$)†  & 0.615 & 2.54$\times$ & 3.75 & 5.70 & 207.57 \\
    \midrule
    \textbf{DDIM-25 steps†}          & 0.818 & 1.91$\times$ & 3.18 & 4.74 & 232.01 \\
    FORA ($\mathcal{N}=3$)†          & 0.701 & 2.23$\times$ & 3.55 & 6.36 & 229.02 \\
    TaylorSeer ($\mathcal{N}=3$)        & 0.764 & 2.05$\times$ & 2.34 & 4.69 & \underline{238.07} \\
    \rowcolor{gray!20}
    \textbf{X-Slim(S)-slow}          & 0.668 & 2.34$\times$ & \textbf{2.32} & \textbf{4.34} & \textbf{242.11} \\
    \rowcolor{gray!20}
    \textbf{X-Slim(C2F)-slow}        & 0.611 & 2.56$\times$ & \underline{2.33} & \underline{4.49} & 237.97 \\
    \midrule
    \textbf{DDIM-20 steps†}          & 0.679 & 2.30$\times$ & 3.81 & 5.15 & 221.43 \\
    FORA ($\mathcal{N}=5$)†          & 0.602 & 2.95$\times$ & 6.58 & 11.29 & 193.01 \\
    TaylorSeer ($\mathcal{N}=5$)†        & 0.663 & 2.36$\times$ & 2.65 & 5.34 & 231.11 \\
    \rowcolor{gray!20}
    \textbf{X-Slim(S)-fast}          & 0.553 & 2.83$\times$ & \textbf{2.38} & \textbf{4.55} & \textbf{236.30}\\
    \rowcolor{gray!20}
    \textbf{X-Slim(C2F)-fast}        & 0.500 & 3.13$\times$ & \underline{2.42} & \underline{4.59} & \underline{235.64} \\
    \bottomrule
\end{tabular}
}
\footnotesize
\begin{itemize}
    \item{†} The Visual Quality results of marked methods are adopted from TaylorSeer\cite{taylorseers}.
\end{itemize}
\vspace{-0.4cm}
\end{table}

{\bf{Text-to-Video Generation.}} On HunyuanVideo, we generate 81-frame videos at 544$\times$960 resolution. 
TaylorSeer runs out of memory due to storing all block features across timesteps. 
We compare X-Slim with TeaCache and EasyCache, using thresholds (0.15, 0.25) and (0.035, 0.055) as baselines.  
Our X-Slim(C2F)-fast achieves 3.52$\times$ acceleration with a strong VBench\cite{vbench} score of 81.69, showing lower LPIPS\cite{lpips} than all baselines. It should be noted that EasyCache's high PSNR/SSIM\cite{iqa} scores are misleading. Its outputs are visibly noisy and blurry, with poor perceptual quality (see Supplementary Section~\ref{sec:suppl-visual-hunyuan}).

\begin{table*}[htbp]
\centering
\caption{\textbf{Quantitative comparison in text-to-video generation }on Vbench\cite{vbench} with HunyuanVideo.}
\vspace{-0.2cm} 
\label{tab_hunyuan}
\setlength\tabcolsep{11.8pt}  
\renewcommand{\arraystretch}{0.9} 

\begin{tabular}{l|cc|cccc}
\toprule
\textbf{Method}
 & \multicolumn{2}{c|}{\textbf{Efficiency}}
 & \multicolumn{4}{c}{\textbf{Visual Quality}} \\
\textbf{HunyuanVideo}
 & \textbf{Latency(s)$\downarrow$} & \textbf{Speed$\uparrow$}
 & \textbf{VBench(\%)$\uparrow$} & \textbf{PSNR$\uparrow$} & \textbf{SSIM$\uparrow$} & \textbf{LPIPS$\downarrow$} \\
\midrule
\textbf{Original: 50 steps}       & 597.92          & 1.00$\times$ & 82.10 {\scriptsize \textcolor{gray}{(+0.00)}}& --     & --     & --     \\
\midrule
$50\%$: 25 steps                 & 301.98          & 1.98$\times$ & 81.68 {\scriptsize \textcolor{gray}{(-0.42)}} & 19.94  & 0.7367 & 0.2673 \\
Uniform Cache ($\mathcal{N}=2$)  & 290.25          & 2.06$\times$ & 81.73 {\scriptsize \textcolor{gray}{(-0.37)}} & 19.82  & 0.7294 & 0.2708 \\
EasyCache($\delta=0.035$)        & 231.75          & 2.58$\times$ & 81.51 {\scriptsize \textcolor{gray}{(-0.59)}} & \textbf{30.56}  & \textbf{0.8977} & \textbf{0.0814} \\
TeaCache ($\delta=0.15$)         & 254.43          & 2.35$\times$ & \underline{81.90} {\scriptsize \textcolor{gray}{(-0.20)}} & 23.18  & 0.8037 & 0.1827 \\
\rowcolor{gray!20}
\textbf{X-Slim(S)-slow}          & 233.56          & 2.56$\times$ & \textbf{81.92} {\scriptsize \textcolor{blue}{(-0.18)}} & \underline{28.46} & \underline{0.8920}  & \underline{0.0893}  \\
\rowcolor{gray!20}
\textbf{X-Slim(C2F)-slow}        & 203.37          & 2.94$\times$ & 81.87 {\scriptsize \textcolor{blue}{(-0.23)}} & 27.52  & 0.8794 & 0.1062  \\
\midrule
$30\%$: 15 steps                 & 179.56          & 3.33$\times$ & 81.30 {\scriptsize \textcolor{gray}{(-0.80)}} & 17.33  & 0.6674 & 0.3692 \\
Uniform Cache ($\mathcal{N}=3$)  & 196.04          & 3.05$\times$ & 81.52 {\scriptsize \textcolor{gray}{(-0.58)}} & 17.61  & 0.6717 & 0.3524 \\
EasyCache($\delta=0.055$)        & 188.62          & 3.17$\times$ & 80.77 {\scriptsize \textcolor{gray}{(-1.33)}} & \textbf{27.39}  & \textbf{0.8357} & 0.1745 \\
TeaCache ($\delta=0.25$)         & 179.12          & 3.34$\times$ & 81.30 {\scriptsize \textcolor{gray}{(-0.80)}} & 23.10  & 0.7946 & 0.1973 \\
\rowcolor{gray!20}
\textbf{X-Slim(S)-fast}          & 183.41          & 3.26$\times$ & \textbf{81.75} {\scriptsize \textcolor{blue}{(-0.35)}} & \underline{24.61}  & \underline{0.8346} & \textbf{0.1619} \\
\rowcolor{gray!20}
\textbf{X-Slim(C2F)-fast}        & 169.86          & 3.52$\times$ & \underline{81.69} {\scriptsize \textcolor{blue}{(-0.41)}} & 24.07 & 0.8216 & \underline{0.1638}  \\
\bottomrule
\end{tabular}
\vspace{-0.4cm}    
\end{table*}

{\bf{Class-Conditional Image Generation.}} Following TaylorSeer, we evaluate X-Slim on DiT-XL/2 against TaylorSeer, ToCa, FORA\cite{fora}, and reduced-step DDIM\cite{ddim}. 
As shown in Tab.~\ref{tab_dit}, X-Slim(C2F)-fast reaches 3.13$\times$ acceleration with an FID-50k\cite{fid} of 2.42, achieving the best trade-off between speed and quality.

\begin{figure*}[!t]
  \centering
  \begin{adjustbox}{center,max width=\textwidth}
    \includegraphics[width=\textwidth]{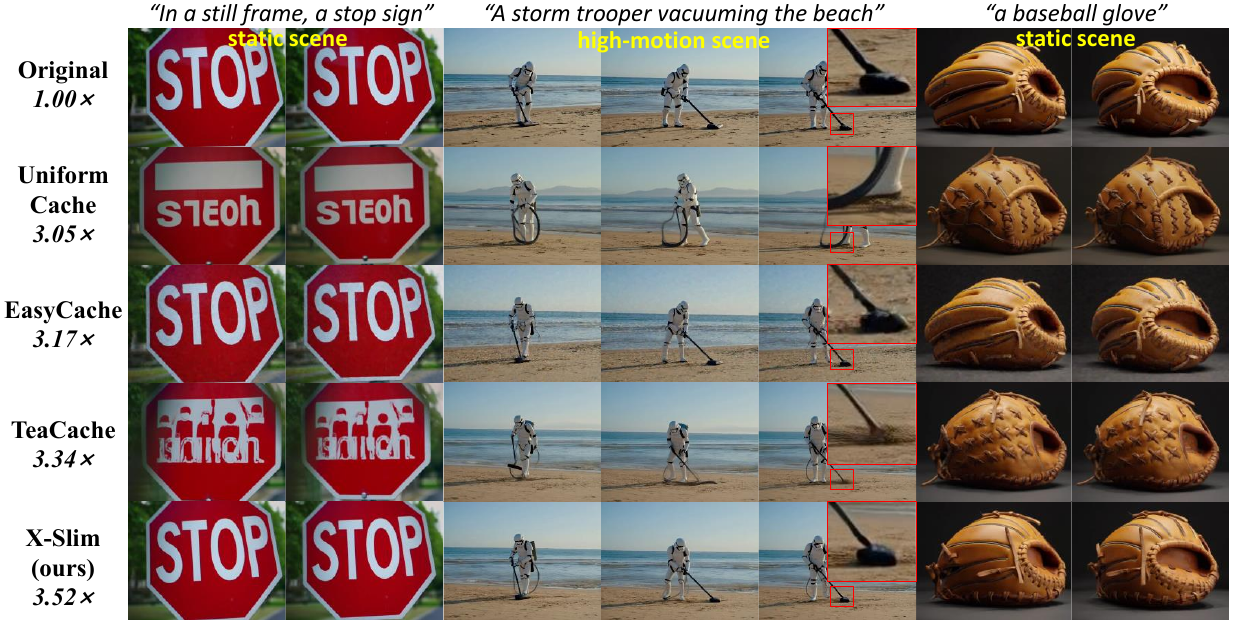}
  \end{adjustbox}
  \caption{Visual quality comparison on HunyuanVideo. Other methods show text errors and quality loss on static scenes and artifacts or content loss on high motion videos. On the high-motion prompt, Uniform Cache shows frame errors, EasyCache is noisy and blurry, and TeaCache loses fine details. Note that EasyCache is structurally close, yet quality degrades in most outputs with visible artifacts. We provide clearer comparisons in Supplementary Section~\ref{sec:suppl-visual-hunyuan}.}

  \vspace{-0.4cm}
  \label{fig:compar-hunyuan}
\end{figure*}

\paragraph{Visualization.}
Figs.~\ref{fig:compar-flux} and \ref{fig:compar-hunyuan} present qualitative results on text-to-image and text-to-video. 
We compare X-Slim with DDIM-50, UniformCache, TeaCache, and TaylorSeer. For videos, we include EasyCache for comparison, while TaylorSeer is omitted due to out-of-memory (OOM) failures.
Across both tasks, X-Slim achieves the best speed–quality trade-off, producing sharper structures and more consistent details than competing methods.

\subsection{Ablation Studies}
We conduct ablations on FLUX.1-dev to answer four questions about \textbf{X-Slim}:

\textbf{Q1) } Why switch to lightweight refresh rather than keep increasing step-level skipping?  
\textbf{Q2) } Why introduce {\bf{both}} block refresh and token refresh?
\textbf{Q3) } How should we set the two thresholds ($\delta_{\text{warn}}$, $\delta_{\text{crit}}$)?  
\textbf{Q4) } What parameter-setting benefits does the push-then-polish paradigm offer?

\begin{figure}[!t]
  \centering
  \includegraphics[width=\linewidth]{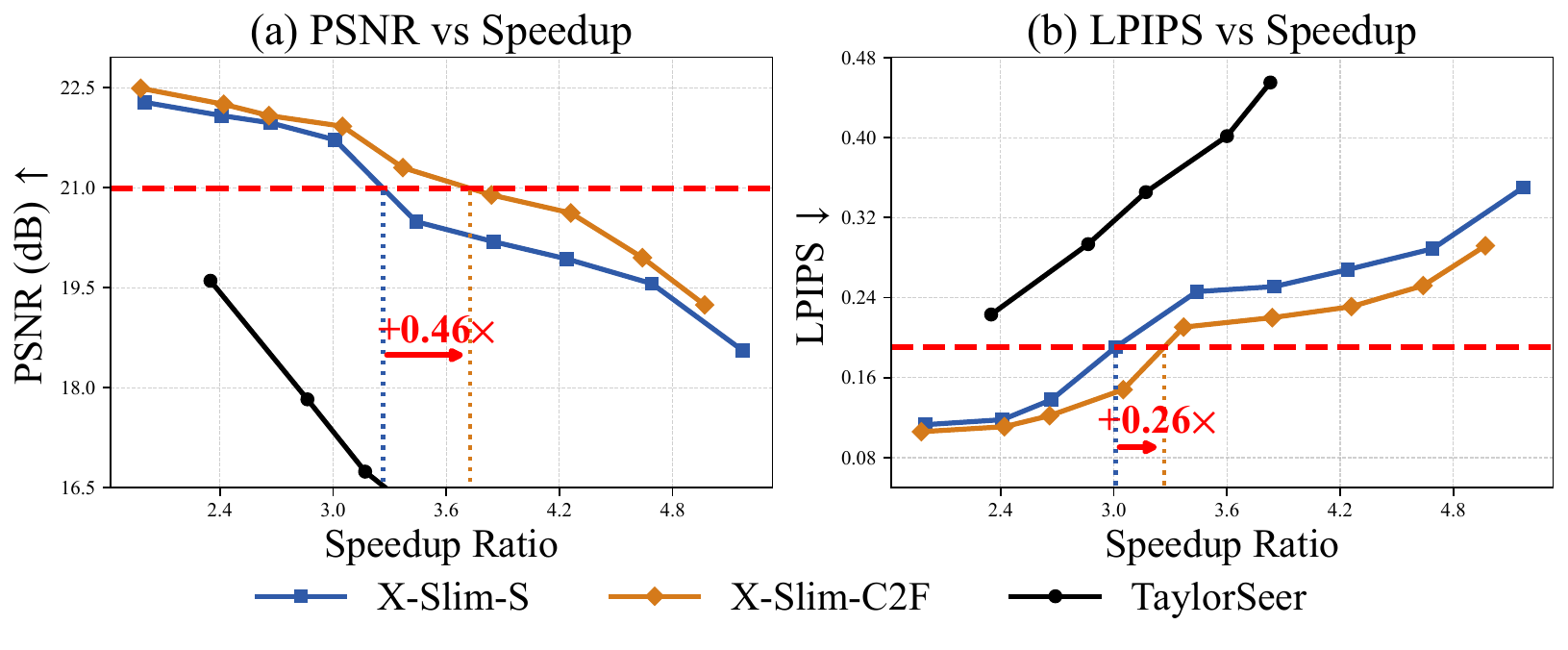}
  \vspace{-0.4cm}
  \caption{Comparison of X\mbox{-}Slim(C2F) and X\mbox{-}Slim(S) at similar speedups. As acceleration increases, the quality gap widens and X\mbox{-}Slim(C2F) clearly outperforms X\mbox{-}Slim(S). }
  \label{fig:ablationA1}
\end{figure}

\paragraph{A1: Limits of Aggressive Step-Level Reuse.}
In Fig.~\ref{fig:ablationA1}, we compare X-Slim(S) and X-Slim(C2F) at similar speedups using PSNR and LPIPS.
At matched speed, X-Slim(C2F) yields better quality than X-Slim(S) consistently. At matched quality, it achieves higher speed.
The reason is that X-Slim(S) removes fully computed steps that ordinarily corrects accumulated global error, causing quality to drop.
In contrast, X-Slim(C2F) retains that global correction and trims its cost by removing only low-impact blocks and tokens, thereby preserving fidelity while maintaining acceleration.

\maketitle

\paragraph{A2: Complementarity of block and token refresh.}
\label{paragraph:A2}

We first reach 3.01$\times$ speedup using X-Slim(S) for the first-stage slimming. From there we compare three options to go faster: +block-only, +token-only, and +block\&token (C2F), with the refresh ratio fixed at 0.8. 
\begin{table}[ht]
  \centering
  \vspace{-0.2cm} 
  \caption{The ablation study for different refresh levels.}
  \vspace{-0.2cm} 
  \label{tab:ablation_a2}
  \setlength{\tabcolsep}{2.2pt}
  \renewcommand{\arraystretch}{0.9}
  \small
  \begin{tabular}{l|cc|ccc}
    \toprule
    \textbf{Method} & \textbf{Latency(s)} & \textbf{Speed}$\uparrow$ & \textbf{PSNR} $\uparrow$ & \textbf{SSIM} $\uparrow$ & \textbf{LPIPS} $\downarrow$ \\
    \midrule
    X\mbox{-}Slim(S) & 9.08  & 3.01$\times$ & 21.72 & 0.82 & 0.19 \\
    \cmidrule(lr){1-6} 
    + block-only     & 8.13  & 3.36$\times$ & 20.64 & 0.76 & 0.27 \\
    + token-only     & 8.08  & 3.38$\times$ & 21.01 & 0.80 & 0.24 \\
    \rowcolor{gray!20}
    \textbf{+ block\&token} & 8.11  & \textbf{$3.37\times$} & \textbf{21.30} & \textbf{0.81} & \textbf{0.21} \\
    \bottomrule
  \end{tabular}
  \vspace{-0.5cm}
\end{table}

\begin{figure}[!t]
  \centering
  \includegraphics[width=\linewidth]{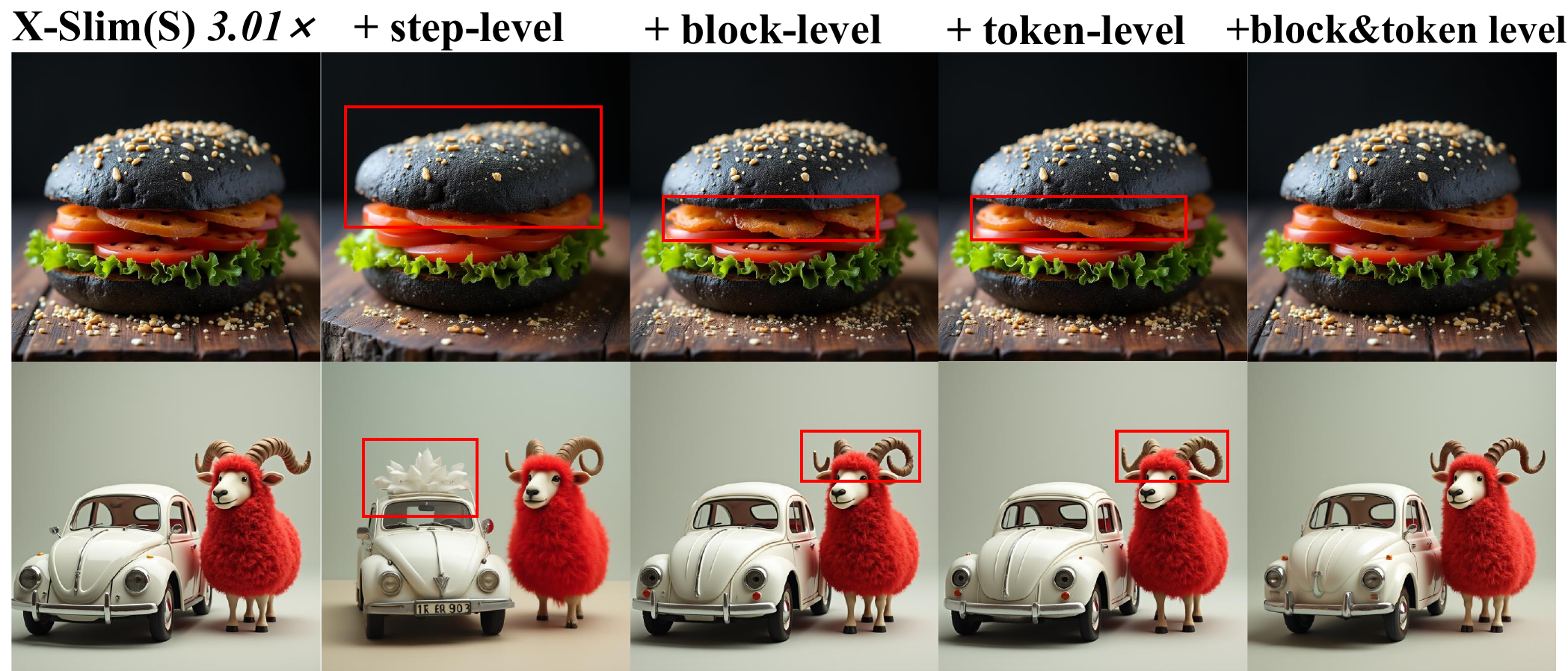}
  \vspace{-0.5cm}
  \caption{Ablation A2 visualization. When pushing beyond the initial 3.01× speedup, the step-only variant shows noticeable structural artifacts. Block- or token-only introduces errors or detail mismatches. The block+token variant yields stronger global consistency and cleaner local details.}
  \vspace{-0.5cm}
  \label{fig:ablationA2}
\end{figure}

As shown in Tab.~\ref{tab:ablation_a2}, at comparable speedups the block+token scheme attains higher quality. 
In Fig.~\ref{fig:ablationA2}, we visualize the effects of using different level settings. At comparable speedups, X\mbox{-}Slim(C2F) outperforms single-level settings with stronger global consistency and cleaner local details.

\paragraph{A3: Selecting $\delta_{\text{warn}}$ and $\delta_{\text{crit}}$ for controlled quality.}
Under a controlled setup with refresh ratio 0.8 and the controller initialized at \(\delta_{\text{warn}}=\delta_{\text{crit}}\) which gives \(3.01\times\) speedup, we tune \(r=\delta_{\text{warn}}/\delta_{\text{crit}}\).
In Fig.~\ref{fig:warn-crit}, when \(r\) is close to \(1.0\), the dual threshold controller becomes ineffective and behaves as if a single threshold were used.
When \(r\) is very small, refresh triggers too early and acceleration drops.
We set \(r\) to the mean relative change on the reuse error mid plateau, which yields a stable trade off with timely refresh before drift and full inference only when necessary.

\paragraph{A4. Simple tuning with no large search space.}
Directly mixing three levels forces per-step choices over the level and reuse ratio, creating a large search space and heavy manual tuning. Even a token-only sweep can exceed 82 GPU hours\cite{astraea}. Our push-then-polish paradigm collapses this space to a few stable knobs and introduces them only on steps that need refresh. When the refresh ratio is below about 0.5, correction is too weak and errors accumulate, which behaves close to no refresh. As the speed target rises there is less slack per step, so a larger refresh ratio is needed to hold quality.
In practice this yields robust performance with minimal tuning and no large grid search.
Further details are provided in Supplementary Section~\ref{sec:parameter-settings}.
\noindent
\begin{minipage}[t]{0.55\linewidth}
  \centering
  \includegraphics[width=\linewidth]{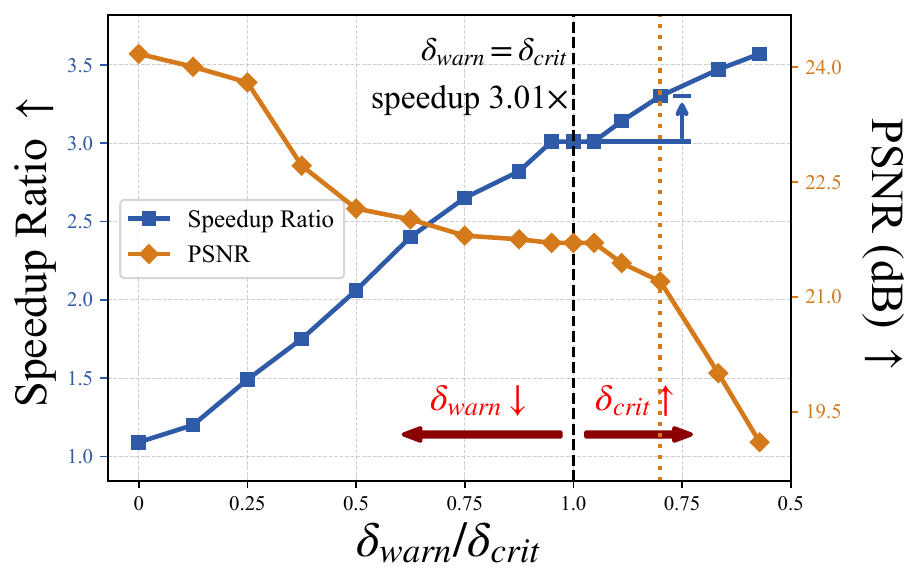}
  \captionof{figure}{Speed and quality curves versus \(r=\delta_{\text{warn}}/\delta_{\text{crit}}\) for ablation A3.}
  \label{fig:warn-crit}
\end{minipage}
\hfill
\begin{minipage}[t]{0.43\linewidth}
  \centering
  \includegraphics[width=\linewidth]{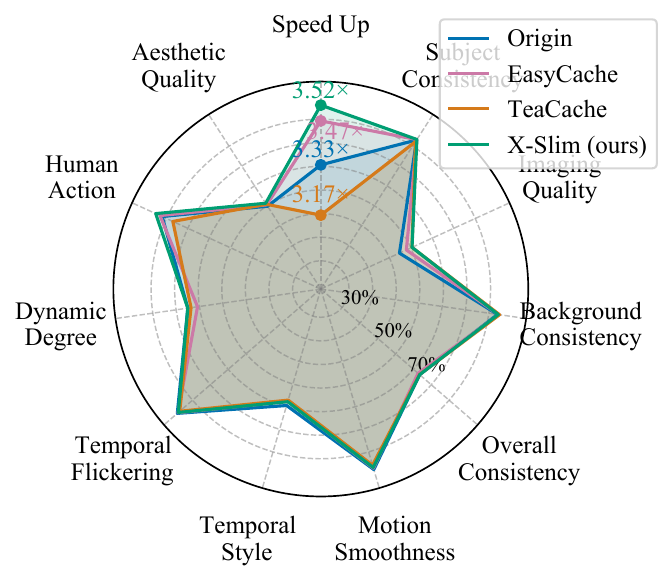}
  \captionof{figure}{Comparison of VBench metrics and acceleration across methods.}
  \label{fig:vbench}
\end{minipage}

\section{Conclusion}
\label{sec:conclu}
In this work, we introduce \textbf{X\text{-}Slim}, a training-free cache-based accelerator that, to our knowledge, is the first unified approach to exploit cacheable redundancy across time, structure, and space, thereby fully harvesting reusable computation. 
X\text{-}Slim instantiates a \emph{push-then-polish} paradigm coupled with a \emph{level-specific} design, where a dual-threshold controller and a hybrid scheme work together to deliver precise per-level decisions at minimal overhead and with no retraining. 
In plug-and-play use, X-Slim consistently outperforms prior accelerators, delivering up to 4.97× and 3.52× lower latency on text-to-image and text-to-video, respectively, while on DiT-XL/2 it achieves a 3.13× speedup together with a 2.42 FID reduction, pushing the frontier of speed and quality.

{
    \small
    \bibliographystyle{ieeenat_fullname}
    \bibliography{main}
}

\clearpage
\setcounter{page}{1}
\maketitlesupplementary

\section{Universal U-Curve Across Prompts}
\label{sec:diff-prompts}

\paragraph{Setup.}
We analyze adjacent timesteps during denoising using prompts from the HPS benchmark across four styles, namely Animation, Concept\mbox{-}art, Painting, and Photo, randomly sampling \textbf{100 prompts per style}. 
For each pair of adjacent steps \(t{-}1\) and \(t\), we compute two inline metrics on hidden states: the relative \(L_1\) change \( \|\Delta_t - \Delta_{t-1}\|_{1} / \|\Delta_{t-1}\|_{1} \), and the cosine similarity between \(\Delta_t\) and \(\Delta_{t-1}\). 
These quantify the magnitude of reuse error and the directional consistency of features.

\begin{figure}[!ht]
    \centering
    \includegraphics[width=\linewidth]{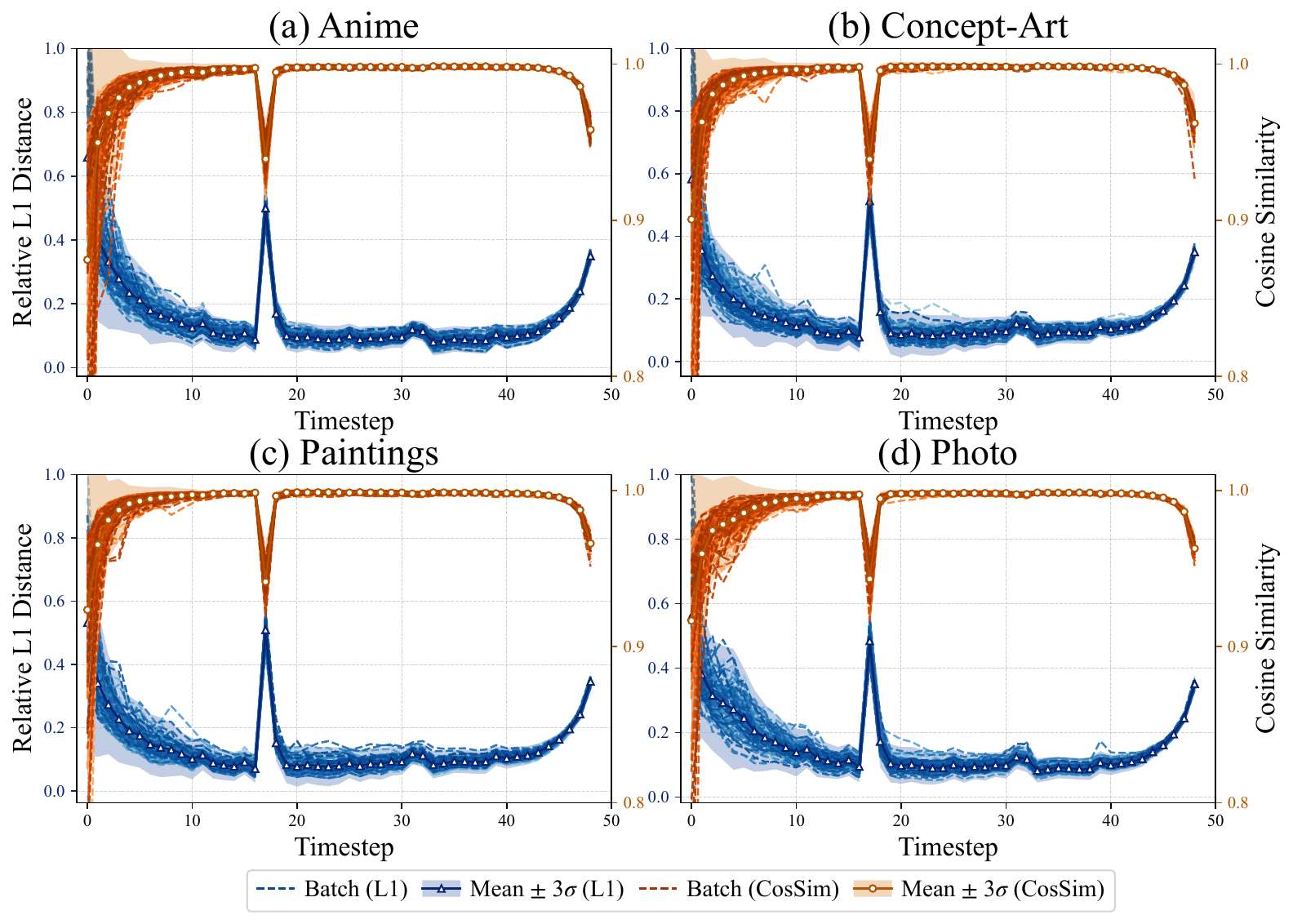}
    \caption{Universal U-Curve Across Prompts.}
    \label{fig:suppl6}
    \vspace{-0.4cm}
\end{figure}
\paragraph{Finding and implication.}
Across styles and prompts, the curves exhibit a clear U\mbox{-}shape: early and late steps show larger relative change and lower cosine similarity, while mid\mbox{-}trajectory steps show smaller change and cosine similarity close to 1.0. 
The pattern is governed primarily by the model's denoising dynamics, with weak prompt influence, suggesting a schedule that allocates more computation to the head and tail and less to the middle. 
We also examine representative dynamic timestep methods such as TeaCache and EasyCache and observe that strategies tend to converge to nearly the same schedule across prompts, differing mainly at the high\mbox{-}variation ends, yet they require nontrivial preprocessing.

\paragraph{Our choice and result.}
Guided by the prompt\mbox{-}robust U\mbox{-}curve, we adopt a \emph{static} step\mbox{-}level schedule calibrated once on a small set, which is simple, training\mbox{-}free, and aligned with the backbone's intrinsic dynamics. 
In the experimental tables of Section~4, under matched speed our schedule attains higher perceptual quality than dynamic timestep policies, and under matched quality it achieves higher speed, confirming that our method is both practical and effective.

\section{Position-Sensitive Reuse Error}
\label{sec:position-sensitivity}

Building on Supplementary Section~\ref{sec:diff-prompts}, the effect of caching is governed by the model's intrinsic denoising dynamics rather than prompt\mbox{-}specific effects.
The \emph{early} and \emph{late} steps undergo large, model\mbox{-}driven changes and are sensitive to perturbations, while the \emph{middle} steps evolve more smoothly and are comparatively stable.
We verify this by injecting caching in three phases (early, middle, late) and inspecting the final images.

\begin{figure}[!ht]
    \centering
    \includegraphics[width=\linewidth]{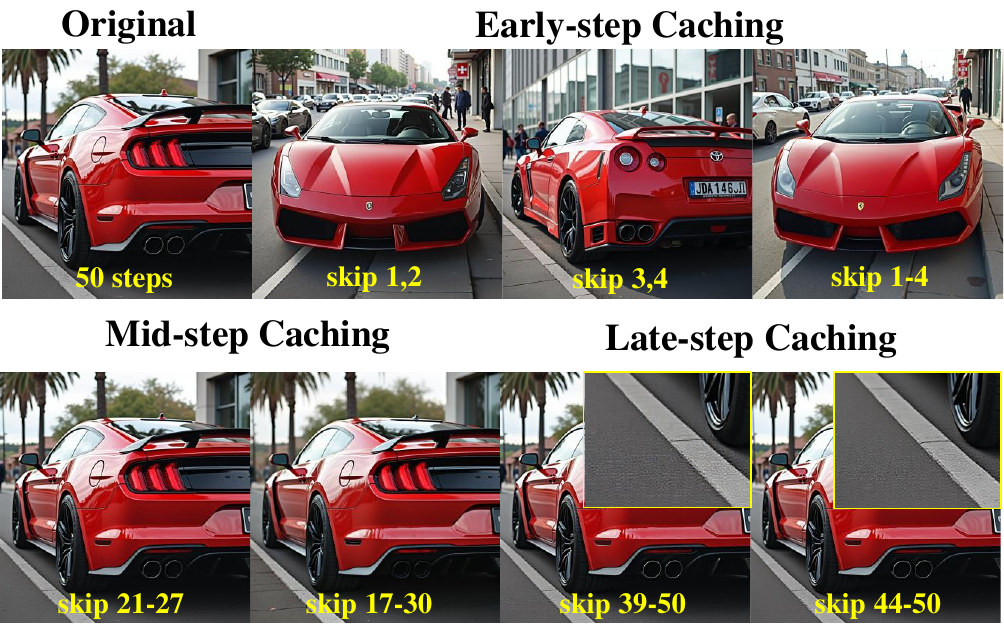}
    \caption{Position-Sensitive Ablations.}    
    \label{fig:suppl7}
    \vspace{-0.2cm}
\end{figure}

As shown in Fig.~\ref{fig:suppl7}, caching in the early phase often distorts global structure and reduces quality because this stage establishes semantic layout and coarse geometry and small errors are amplified.
Caching in the late phase preserves the layout but hurts fidelity: details blur and residual noise appears since this stage refines high\mbox{-}frequency content.
By contrast, middle\mbox{-}phase caching is consistently robust: even with stronger reuse, both structure and perceptual quality remain close to a full run.

These observations yield a simple rule that follows the model's behavior: use less caching at the fragile head and tail, and more caching in the stable middle.
X\text{-}Slim encodes this with its push\mbox{-}then\mbox{-}polish controller.
Thresholds limit caching near the boundaries by triggering refresh or full steps when needed, while the middle region gains larger reuse to capture most of the speedup without sacrificing quality.

\section{Detailed Experimental Settings}
\label{sec:detail-exp-settings}
\paragraph{Model Configurations.}
We follow TaylorSeer's setup and evaluate three diffusion backbones: FLUX.1-dev for text-to-image, HunyuanVideo for text-to-video, and DiT-XL/2 for class-conditional generation.
For FLUX.1-dev, we use 200 DrawBench prompts at $1024\times 1024$ and sample 5 images per prompt (1{,}000 images total).
For HunyuanVideo, we adopt the VBench suite with 946 prompts and generate five videos per prompt, totaling 4{,}730 clips. We use 540p with 81 frames for batch quality and latency evaluation, and 480p for visualization.
For DiT-XL/2 on ImageNet, we uniformly cover the 1{,}000 classes and produce 50{,}000 images at $256\times 256$.

\paragraph{Evaluation and Metrics.}
We assess efficiency by inference latency and quality by task-appropriate metrics.
For text-to-image, we report ImageReward and CLIP for perceptual quality and text–image alignment, and also measure LPIPS, PSNR, and SSIM against full-compute references.
For text-to-video, we follow VBench metrics.
For class-conditional generation, we report FID-50k as the primary metric, complemented by sFID and Inception Score to capture fidelity and diversity.

\paragraph{Hardware and Computational Resources.}
All images and videos used for quality evaluation are generated in data parallel on 32 Ascend 910B devices.
Latency is measured on a single A100 GPU to ensure a fair comparison.

\section{Refresh settings and ratio selection.}
\label{sec:parameter-settings}

\paragraph{Setup.}
We evaluate the impact of refresh ratios on FLUX.1-dev under two runtime modes named X\text{-}Slim\text{-}slow and X\text{-}Slim\text{-}fast. The dual-threshold ratio is set to \(r=\delta_{\text{warn}}/\delta_{\text{crit}}=0.9\). When \(r=1\) the policy reduces to the step only variant X\text{-}Slim(S). Block refresh is controlled by \(\delta_{\text{blk}}\) which determines the effective refresh ratio. We ablate refresh ratios at 1.0, 0.8, 0.5, 0.2, and 0. Results are reported in Tables~\ref{tab:suppl_a3_speed3} and \ref{tab:suppl_a3_speed5}.

\paragraph{Key findings.}
Starting from our step-only variant X\text{-}Slim(S) at about \(3.01\times\) acceleration, adding a 50\% block refresh and an 80\% token refresh raises the speed to about \(3.37\times\) with negligible quality loss. Lowering the refresh ratios further causes noticeable degradation because correction becomes too weak and errors accumulate, which behaves close to no refresh. When targeting higher acceleration around \(4.69\times\) a larger refresh ratio is needed to maintain quality because fewer timesteps leave less slack per step. Using 80\% for both block and token refresh is the most stable choice in this regime.

\begin{table}[ht]
\centering
\vspace{-0.2cm} 
\caption{Ablation A4 on X\text{-}Slim\text{-}slow refresh ratio effects.}
\vspace{-0.2cm} 
\label{tab:suppl_a3_speed3}
\setlength{\tabcolsep}{2.2pt}
\renewcommand{\arraystretch}{1.1}
\small
\begin{tabular}{l|cc|ccc}
    \toprule
    \textbf{Method} & \textbf{Latency(s)} & \textbf{Speed}$\uparrow$ & \textbf{PSNR} $\uparrow$ & \textbf{SSIM} $\uparrow$ & \textbf{LPIPS} $\downarrow$ \\
    \midrule
    L1.0+Tok1.0  & 9.076 & 3.01 & 21.7168 & 0.8162 & 0.1905 \\
    \midrule
    L0.8+Tok0.8  & 8.618 & 3.17 & 21.4582 & 0.8139 & 0.1950 \\
    L0.8+Tok0.5  & 8.179 & 3.34 & 21.3494 & 0.8102 & 0.2073 \\
    \rowcolor{gray!20}
    \textbf{L0.5+Tok0.8}  & \textbf{8.107} & \textbf{3.37} & \textbf{21.2998} & \textbf{0.8097} & \textbf{0.2105} \\
    \midrule
    L0.5+Tok0.5  & 7.610 & 3.59 & 20.3400 & 0.7849 & 0.2693 \\
    L0.5+Tok0.2  & 7.384 & 3.70 & 20.1050 & 0.7577 & 0.2801 \\
    L0.2+Tok0.5  & 7.324 & 3.73 & 19.9076 & 0.7568 & 0.2819 \\
    L0.2+Tok0.2  & 6.969 & 3.92 & 19.8064 & 0.7456 & 0.3101 \\
    L0.0+Tok0.0  & 6.413 & 4.26 & 19.7784 & 0.7255 & 0.3442 \\
    \bottomrule
\end{tabular}
\vspace{-0.4cm}
\end{table}

\begin{table}[ht]
\centering
\vspace{-0.2cm} 
\caption{Ablation A4 on X\text{-}Slim\text{-}fast refresh ratio effects.}
\vspace{-0.2cm} 
\label{tab:suppl_a3_speed5}
\setlength{\tabcolsep}{2.2pt}
\renewcommand{\arraystretch}{1.1}
\small
\begin{tabular}{l|cc|ccc}
    \toprule
    \textbf{Method} & \textbf{Latency(s)} & \textbf{Speed}$\uparrow$ & \textbf{PSNR} $\uparrow$ & \textbf{SSIM} $\uparrow$ & \textbf{LPIPS} $\downarrow$ \\
    \midrule
    L1.0+Tok1.0  & 5.825 & 4.69 & 19.5601 & 0.7520 & 0.2884 \\
    \midrule
    \rowcolor{gray!20}
    \textbf{L0.8+Tok0.8}  & \textbf{5.497} & \textbf{4.97} & \textbf{19.2407} & \textbf{0.7321} & \textbf{0.2919} \\
    \midrule
    L0.8+Tok0.5  & 5.244 & 5.21 & 18.7047 & 0.7137 & 0.3385 \\
    L0.5+Tok0.8  & 5.214 & 5.24 & 18.5174 & 0.7068 & 0.3402 \\
    L0.5+Tok0.5  & 4.923 & 5.55 & 18.3918 & 0.7042 & 0.3623 \\
    L0.0+Tok0.0  & 3.977 & 6.87 & 18.0763 & 0.6468 & 0.4732 \\
    \bottomrule
\end{tabular}
\vspace{-0.4cm}
\end{table}

\paragraph{Parameter setting and tuning efficiency.}
Directly mixing three levels requires per step decisions on level and refresh ratio which creates a large search space and heavy manual tuning. The push then polish paradigm narrows this to a few stable knobs that are introduced only on steps that require refresh. We tune a single threshold ratio \(r\) by setting it to the average relative change on the mid plateau of the reuse error curve. We set the block and token threshold to match a target refresh rate measured on a small calibration set. In practice the settings fall within a narrow band, which simplifies tuning and improves robustness across backbones.

\section{Visualization on FLUX.1-dev}
\label{sec:suppl-visual-flux}
\begin{figure*}[!t]
  \centering
  \begin{adjustbox}{center,max width=\textwidth}
    \includegraphics[width=1\textwidth]{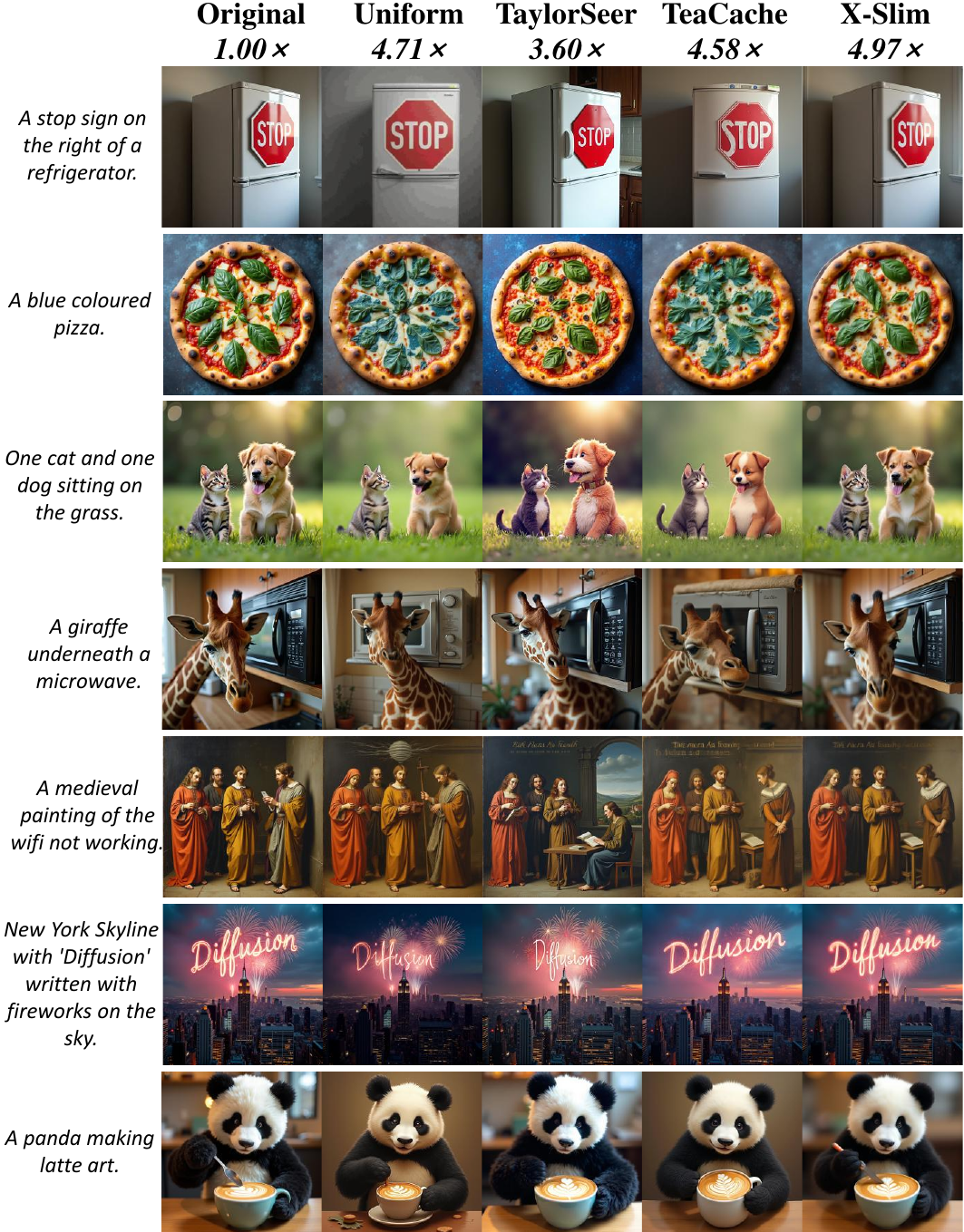}
  \end{adjustbox}
  \vspace{-0.5cm}
  \caption{Qualitative comparison on FLUX.1-dev (1/2).}

  \label{fig:suppl_flux1}
\end{figure*}

\begin{figure*}[!t]
    \centering
    \begin{adjustbox}{center,max width=\textwidth}
      \includegraphics[width=1\textwidth]{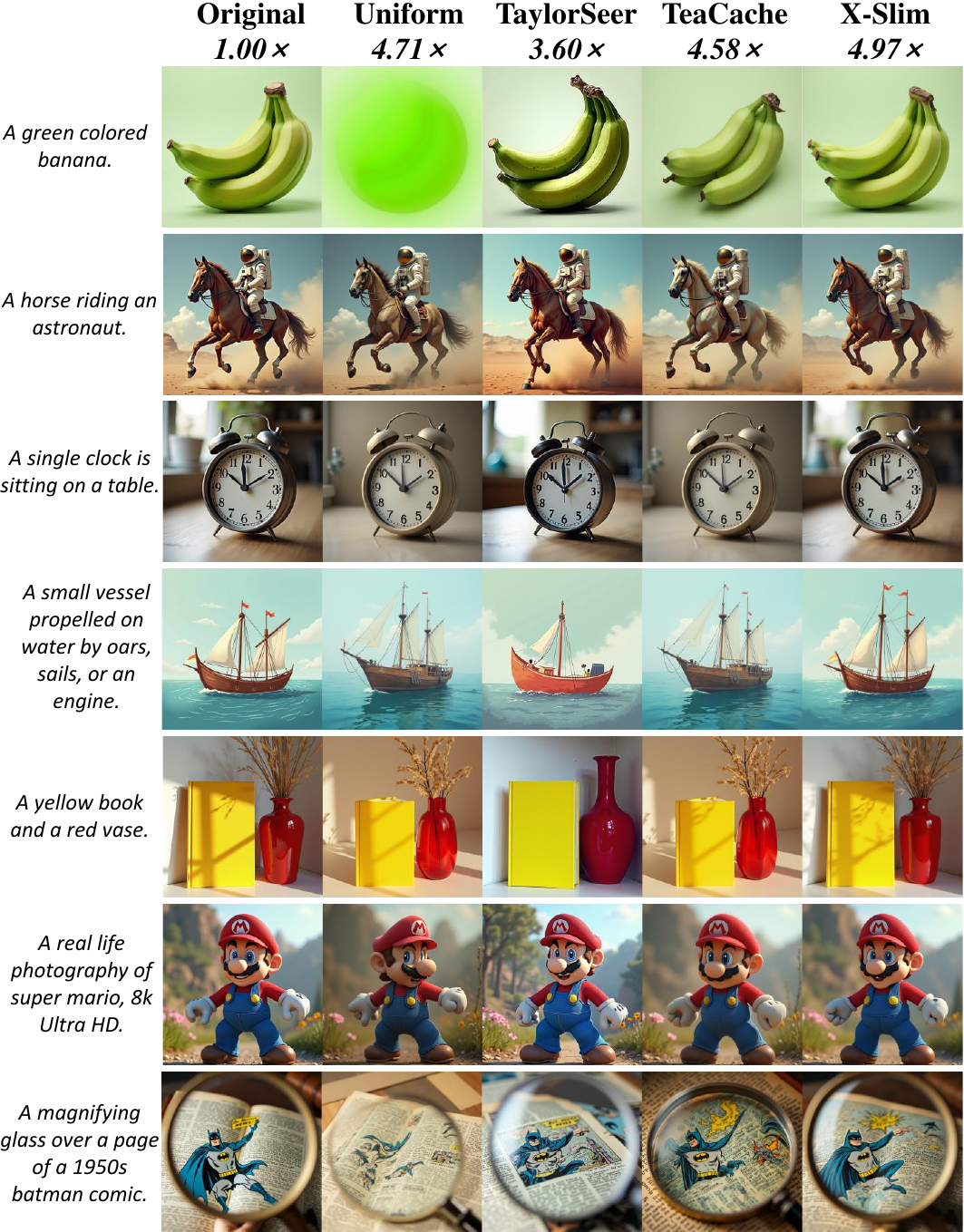}
    \end{adjustbox}
    \vspace{-0.5cm}
    \caption{Qualitative comparison on FLUX.1-dev (2/2).}
  
    \label{fig:suppl_flux2}
  \end{figure*}

We present visual comparisons on FLUX.1-dev. In Figs.~\ref{fig:suppl_flux1} and \ref{fig:suppl_flux2}, X\text{-}Slim reaches \(4.97\times\) acceleration while closely matching the full model, achieving the best balance of speed and fidelity.

\maketitle

\section{Visualization on HunyuanVideo}
\label{sec:suppl-visual-hunyuan}
\begin{figure*}[!t]
  \centering
  \begin{adjustbox}{center,max width=\textwidth}
    \includegraphics[width=1\textwidth]{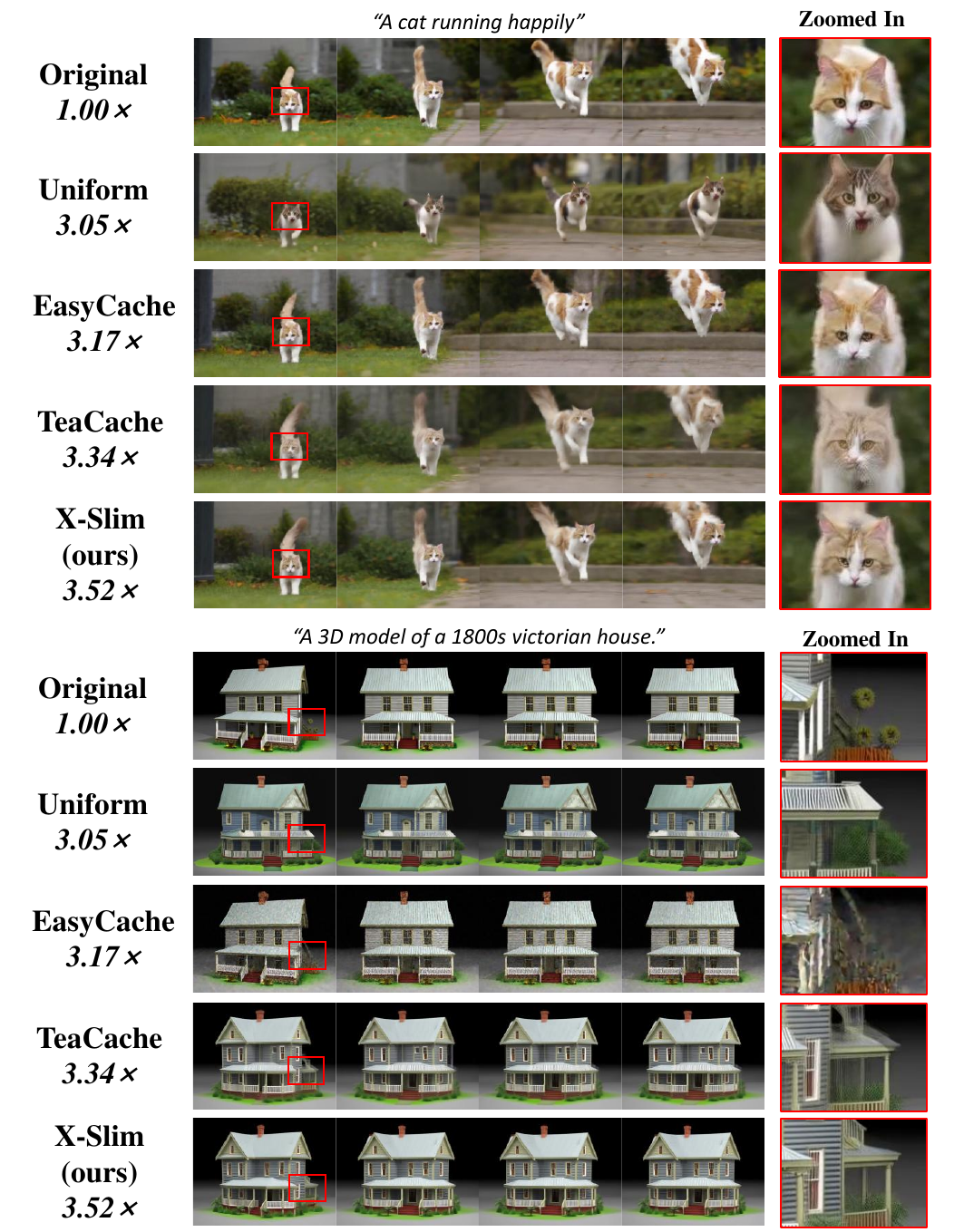}
  \end{adjustbox}
  \vspace{-0.5cm}
  \caption{Qualitative comparison on HunyuanVide (1/2).}
  \label{fig:suppl_hunyuan1}
\end{figure*}

\begin{figure*}[!t]
    \centering
    \begin{adjustbox}{center,max width=\textwidth}
      \includegraphics[width=1\textwidth]{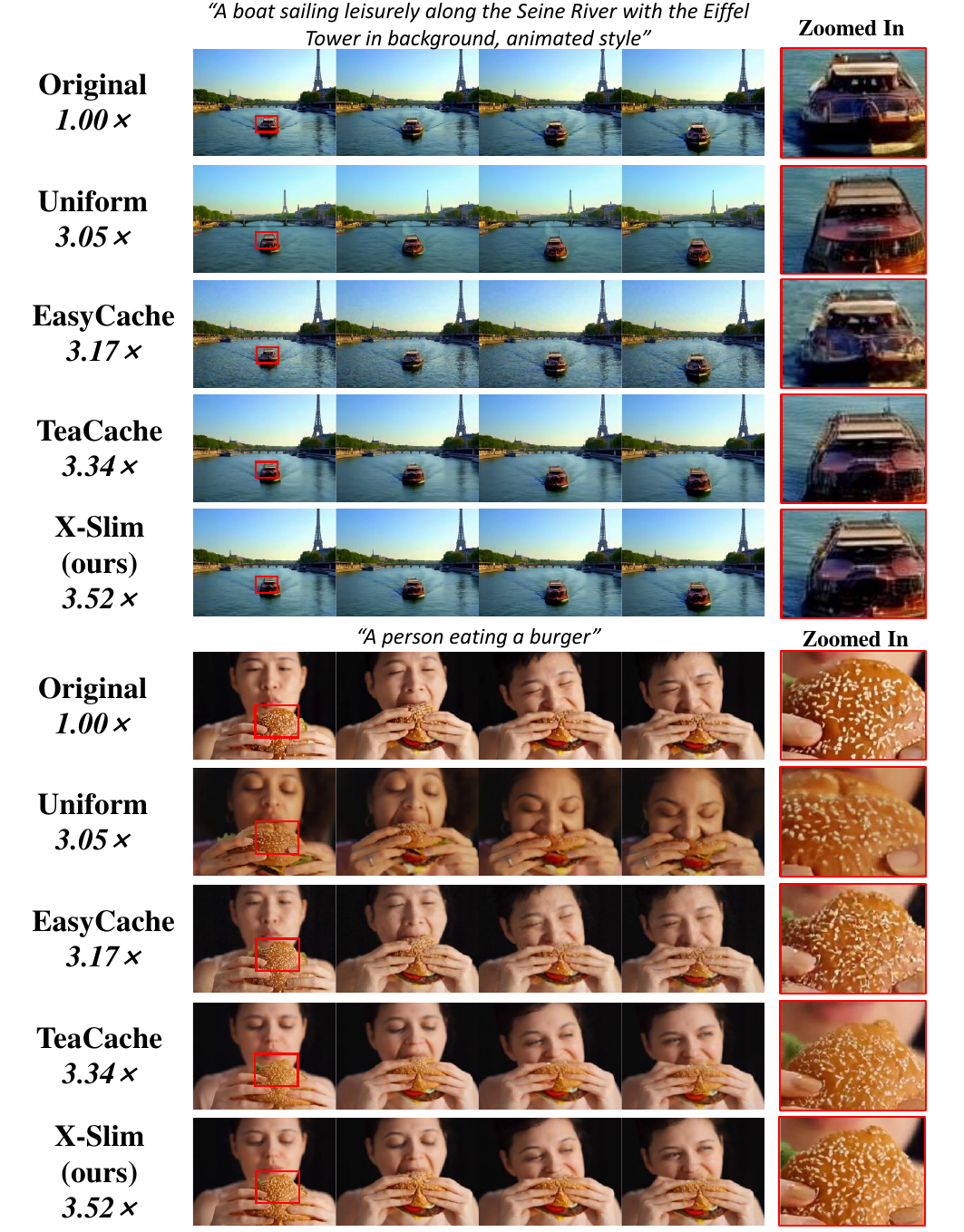}
    \end{adjustbox}
    \vspace{-0.5cm}
    \caption{Qualitative comparison on HunyuanVideo (2/2).}
    \label{fig:suppl_hunyuan2}
  \end{figure*}

As shown in Figs.~\ref{fig:suppl_hunyuan1} and \ref{fig:suppl_hunyuan2}, we provide visual comparisons on HunyuanVideo. At the highest acceleration, X\text{-}Slim produces videos that remain clear and closely match the full model, offering the best balance of speed and fidelity among the compared methods.

\maketitle

\end{document}